\documentclass{article}

\PassOptionsToPackage{numbers, compress}{natbib}
 \usepackage[final]{nips_2016}

\usepackage[utf8]{inputenc} 
\usepackage[T1]{fontenc}    
\usepackage{url}            
\usepackage{booktabs}       
\usepackage{amsfonts}       
\usepackage{nicefrac}       
\usepackage{microtype}      
\usepackage{amssymb}
\usepackage{amsmath}  
\usepackage{bm}
\usepackage{commath}
\usepackage{color}
\usepackage{lineno}
\usepackage[ruled,vlined,linesnumbered]{algorithm2e}
\usepackage{graphicx}
\usepackage{subcaption}
\usepackage{booktabs}
\usepackage{multirow}
\usepackage{tabularx}
\usepackage[flushleft]{threeparttable}
\usepackage[utf8]{inputenc}
\usepackage[english]{babel}
\usepackage{amsthm} 
\usepackage{float}

\newtheorem{theorem}{Theorem}
\newtheorem{remark}{Remark}

\graphicspath{ {Figures/} }

\title{Stochastic Configuration Networks Ensemble\\ for Large-Scale Data Analytics}

\author{
  Dianhui~Wang\thanks{Corresponding author.}, ~~Caihao Cui \\ \\
  Department of Computer Science and Information Technology\\ 
  La Trobe University, 
  Melbourne, VIC 3086, Australia \\
  \texttt{Email:dh.wang@latrobe.edu.au}
}

\begin{document}
\maketitle
\begin{abstract}
This paper presents a fast decorrelated neuro-ensemble with heterogeneous features for large-scale data analytics, where stochastic configuration networks (SCNs) are employed as base learner models and the well-known negative correlation learning (NCL) strategy is adopted to evaluate the output weights.
By feeding a large number of samples into the SCN base models, we obtain a huge sized linear equation system which is difficult to be solved by means of computing a pseudo-inverse used in the least squares method.
Based on the group of heterogeneous features, the block Jacobi and Gauss-Seidel methods are employed to iteratively evaluate the output weights, and a convergence analysis is given with a demonstration on the uniqueness of these iterative solutions.
Experiments with comparisons on two large-scale datasets are carried out,
and the system robustness with respect to the regularizing factor used in NCL is given.
Results indicate that the proposed ensemble learning techniques have good potential for resolving large-scale data modelling problems.  
\end{abstract}

\section{Introduction}
Machine learning has received considerable attention over the past years due to its significant role for data analytics \cite{jordan2015machine}.  
Under big data setting with decentralized information structure, advanced machine learning algorithms with robust and parallel implementations are needed along with the growth of data \cite{scardapane2015distributed, scardapane2016decentralized}.  
Various ensemble learning frameworks, aiming to improve the generalization performance of a learning system, have been developed over the last two decades, and many interesting ideas and theoretical works, including bagging, boosting, AdaBoost and random forests
can be found in \cite{alhamdoosh2014fast, breiman1996bagging, breiman2001random, brown2005managing, chen2009regularized, cui2016high, geman1992neural, hansen1990neural, igelnik1999ensemble, li2015multisource, liu1999ensemble, polikar2012ensemble, rosen1996ensemble, schapire1990strength}.
Generally speaking, learning-based ensembles share some common nature in system design, such as data sampling and the output integration. 
The basis of ensemble learning theory lies in a rational sampling implementation for building each base learner model, which may provide a sound predictability though learning a subset of the whole data set.

For neural network ensembles \cite{chen2009regularized,  igelnik1999ensemble, liu1999ensemble, rosen1996ensemble}, the base models are trained by the error back-propagation (BP) algorithm and the regularizing factor used in the negative correlated cost function can be determined by the cross-validation method. 
Unfortunately, BP algorithm suffers from the sensitive setting of the learning rate, local minima and very slow convergence. 
Therefore, it is challenging to apply the existing ensemble methods for large-scale data sets. 
To overcome this problem,  we employed random vector functional-link (RVFL) networks \cite{igelnik1995stochastic, pao1992functional} to develop a fast decorrelated neuro-ensemble (termed DNNE) in \cite{alhamdoosh2014fast}.
From our experience, DNNE can perform well on smaller data sets \cite{alhamdoosh2014fast,  li2015multisource}.  
However, it is quite limited for dealing with large scale data because of its high computational complexity, the scalability of numerical algorithms for the least squares solution, and hardware constraint (here mainly referring to the PC memory). 
Recall that physical data may come from different types of sensors, localized information source or potential features extracted from multiple runs of some certain feature selection algorithms \cite{cui2016high, hinton2006reducing, lopez2013randomized, nguyen2014multivariate, peng2005feature, reshef2011detecting}.
Thus, for large-scale data analytics, it is useful and significant to develop a generalized neuro-ensemble framework with heterogeneous features. 

This paper is built on our previous work reported in \cite{alhamdoosh2014fast}, which is a specific implementation of the well-known NCL learning scheme using RVFL networks with a default scope setting of the random weights and biases. 
From theoretical statements on the universal approximation property in \cite{igelnik1995stochastic} and our empirical results on RVFL networks in \cite{li2017insights}, the default scope setting (i.e., [-1, 1]) for the random weights and biases cannot ensure the modelling performance at all. 
Therefore, readers should be aware of this pitfall and must be careful in making use of our code\footnote{http://homepage.cs.latrobe.edu.au/dwang/html/DNNEweb/index.html}. Limits of DNNE mainly come from the following aspects: 
(i) the system inputs are centralized or combined with different types of features; 
and (ii) the analysed method of computing the output weights becomes infeasible for large-scale data sets, which is related to the nature of the base learner model (i.e., the number of nodes at the hidden layer must be sufficiently large to achieve sound performance). 
To relax these constraints and emphasize on the fast building of neuro-ensembles with heterogeneous features, we generalize the classical NCL-based ensemble framework into a more general form, where a set of input features are feed into the SCN base models separately. 
This work also provides a feasible solution by using two iterative methods for evaluating the output weights of the SCN ensemble (SCNE). 
In addition, some analyses and discussions on the convergence of these iterative schemes are given through a demonstration on the correlations among the iterative solutions and the pseudo-inverse solution. 

The remainder of the paper is organized as follows: 
Section 2 provides  some technical supports, including the basics of the SCN model, a generalized version of the ensemble generalization error and the negative correlation learning scheme. 
Section 3 describes the proposed SCNE with heterogeneous features, details two iterative learning algorithms and discusses their convergence. 
Section 4 reports some experimental results on two large-scale data sets, including a robustness analysis on the system performance with respect to the regularizing factor used in NCL. 
Section 5 concludes this paper with some remarks on further studies. 

\section{Technical supports}
This section briefly reviews the stochastic configuration networks, extends the ensemble generalization error with heterogeneous features, followed by the negative correlation learning scheme for building ensemble models. 
\subsection{Revisit of stochastic configuration networks}
SCNs are a class of randomized learner models which are recently developed in \cite{WangandLi_SCN}.  
The unique characteristics of the SCN model, different from the classical randomized learner model (i.e., RVFL networks), is the way of generating the random input weights and biases.  
In contrast to RVFL networks, SCNs are built incrementally according to a supervisory mechanism, which constrains the random input weights and biases to take values in a data-dependent territory, namely stochastic configuration support (SCS). 
This constructive approach for building SCNs guarantees the universal approximation property of the resulting SCN model for a given nonlinear map. 
For the sake of completeness, we revisit the main theoretical result in Theorem 1 below. 
 
Given a target function $f: \mathcal{R}^d \rightarrow \mathcal{R}^m$.
Suppose that an SCN model has already been built with $L-1$ hidden nodes, i.e., $f_{L-1} =  \sum_{l=1}^{L-1}\beta_{l}\phi_l(\bm{w}_l^T\bm{x} + b_l)$  ($L=1,2,\dots$; $f_0 = 0$), where $\beta_{l} = [\beta_{l,1}, \beta_{l,2}, \dots, \beta_{l,m}]^T$, and $\phi_l(\bm{w}_l^T\bm{x} + b_l)$ is an activation function of the $l$-th hidden node with random input weights $\bm{w}_l$ and bias $b_l$.
Denoted the residual error by $e_{L-1}^* = f - f_{L-1} = [e_{L-1,1}^*, \dots, e_{L-1,m}^*]$,  where $[\beta_1^{*}, \beta_2^{*},\ldots,\beta_{L-1}^{*}]=\arg \min_{\beta}\|f-\sum_{l=1}^{L-1}\beta_l \phi_l\|$.

Let $\Gamma=\{\phi_1, \phi_2, \phi_3,...\}$ be a set of real-valued functions, and span$(\Gamma)$ denote a function space spanned by $\Gamma$; $L_{2}(D)$ denote the space of all Lebesgue measurable functions $f=[f_1,f_2,\ldots,f_m]:\mathcal{R}^{d}\rightarrow \mathcal{R}^{m}$ defined on $D\subset \mathcal{R}^{d}$, with the $L_2$ norm defined as
\begin{equation}\label{multiple_lp}
  \|f\|=\left(\sum_{q=1}^{m}\int_{D}|f_q(x)|^2dx\right)^{1/2}<\infty.
\end{equation}
The inner product of $\theta=[\theta_1,\theta_2,\ldots,\theta_m]:\mathcal{R}^{d}\rightarrow \mathcal{R}^{m}$ and $f$ is defined as
\begin{equation}\label{multiple_inner}
  \langle f,\theta\rangle=\sum_{q=1}^{m}\langle f_q,\theta_q\rangle=\sum_{q=1}^{m}\int_{D}f_q(x)\theta_q(x)dx.
\end{equation}
\begin{theorem}[\textbf{Wang and Li \cite{WangandLi_SCN}}] 
\label{TH01}
Suppose that span($\Gamma$) is dense in $L_2$ space and for any $\phi \in \Gamma$,  $0 <\| \phi \|<b_{\phi} $ for some $b_{\phi}\in \mathcal{R}^{+}$. Given $0<r<1$ and a nonnegative real number sequence $\{\mu_L\}$ with $\lim_{L\rightarrow+\infty}\mu_L=0$ subjected to $\mu_L\leq (1-r)$.  
For $L=1,2,\ldots$, denoted by
\begin{equation}
\delta_{L}^{*}=\sum_{q=1}^{m}\delta_{L,q}^{*},\quad \delta_{L,q}^{*}=(1-r-\mu_L)\|e_{L-1,q}^{*}\|^2,\quad q=1,2,...,m.
\end{equation}
If the random basis function $\phi_L$ is generated to satisfy the following inequalities:
\begin{equation}
\langle e_{L-1,q}^{*},\phi_L\rangle^2\geq b_{\phi}^2\delta_{L,q}^{*}, \quad q=1,2,...,m,
\end{equation}
and the output weights are evaluated by
\begin{equation}
[\beta_1^{*}, \beta_2^{*},\ldots,\beta_{L}^{*}]=\arg \min_{\beta}\|f-\sum_{l=1}^{L}\beta_l \phi_l\|.
\end{equation}
Then, we have $\lim_{L\rightarrow +\infty}\|f-f_L^{*}\|=0,$ where $f_L^{*}=\sum_{l=1}^{L}\beta_{l}^{*}\phi_l$, $\beta_{l}^{*}=[\beta^{*}_{l,1},\beta^{*}_{l,2},\ldots,\beta^{*}_{l,m}]^{T}$.
\end{theorem} 
Given a training data set with $N$ sample pairs $\{(\bm{x}_n, \bm{y}_n)$, $n=1,2,\ldots,N\}$, where $\bm{x}_n \in \mathcal{R}^d$ and $\bm{y}_n \in \mathcal{R}^m$. 
Let $X \in \mathcal{R}^{N\times d}$ and $Y \in \mathcal{R}^{N\times m}$ represent the input and output data matrix, respectively;  
$e_{L-1}(X) \in \mathcal{R}^{N\times m}$ be the residual error matrix,  where each column $e_{L-1,q}(X)=[e_{L-1,q}(\bm{x}_1),\ldots,e_{L-1,q}(\bm{x}_N)]^T\in \mathcal{R}^N$, $q=1,2,\ldots,m$. 
Denote the output vector of the $L$-th hidden node $\phi_L$ for the input $X$ by 
\begin{equation}
h_L(X)=[\phi_{L}(\bm{w}_L^Tx_1+b_L),\ldots,\phi_{L}(\bm{w}_L^Tx_N+b_L)]^T.
\end{equation}
Thus, the hidden layer output matrix of $f_L$ can be expressed as $H_L=[h_1,h_2,\ldots,h_L]$. Denoted by 
\begin{equation}
\xi_{L,q} = \frac{\Big(e^{T}_{L-1,q}(X)\cdot h_L(X)\Big)^2}{h^{T}_L(X)\cdot h_L(X)} -(1-r-\mu_L)e^{T}_{L-1,q}(X)e_{L-1,q}(X), \quad q = 1,2,\dots, m.
\end{equation}
The SC-III algorithm reported in \cite{WangandLi_SCN} firstly generates a large pool of $T_{max}$ candidate nodes, namely $\{\phi_L^{(1)}, \phi_L^{(2)},\dots, \phi_L^{(T_{max})}\} $, in varying intervals.  
Then, it picks up those candidate nodes whose minimal value of the set $\{\xi_{L,1}, \dots, \xi_{L,m} \}$ is positive.
Then, the candidate node $\phi_L^*$ with the largest value of $\xi_L = \sum_{q=1}^{m}\xi_{L,q}$ will be assigned as the $L$-th hidden node for $f_L$.
Thus, the output weight matrix of the SCN model, $\bm{\beta} = [\beta_1,\beta_2,  \dots, \beta_L]^T \in \mathcal{R}^{L\times m}$, could be computed by the standard least squares method, that is, 
\begin{equation}
  \bm{\beta}^{*}=\arg\min_{\bm{\beta}}\|H_L \bm{\beta}-Y\|_{F}^2=H^{\dagger}_LY,
\end{equation}
where $H^{\dagger}_L$ is the Moore-Penrose generalized inverse of the matrix $H_L$, and $\|\cdot\|_{F}$ represents the Frobenius norm \cite{golub1996Matrix}.

If there is no candidate node that satisfies these conditions, the SC algorithm will automatically increase the value of $r$ to relax the constraints and further adjust the scoping interval to generate a new pool of candidate nodes. This process continues until the residual error decreases to a predefined tolerance $\tau$. To speed up the procedure of building SCN models, we could add the top $n_B$ ranked (according to the values of $\xi_L$) candidate nodes as a batch in each incremental loop of the SC algorithm. With a proper  setting on $n_B$, the batch version of SC-III algorithm could greatly shorten the training time without weakening the prediction performance of the SCN model.
Readers could find more details about the proof of the theorem and the SC algorithms in \cite{WangandLi_SCN}.

\subsection{Generalization error of ensembles with heterogeneous features} 
For data regression, an ensemble model $\bar{f}$ that approximates an unknown target function $g$ could be accomplished by fitting a collection of training samples $D_t = \{(\bm{x}_n,y_n),n=1,2,\dots,N\} $, where $\bm{x}_n  \in \mathcal{R}^{d}$, $y_n \in \mathcal{R}$, and the sample pair $(\bm{x}_n, y_n)$ is an independent and identically distributed (i.i.d) sample from an unknown joint probability distribution $p(\bm{x}, y)$.
Note that the training set $D_t$ is a realization of a random sequence $D$ that shares the same distribution $p(\bm{x}, y)$.

Suppose that the input $\bm{x}$ is concatenated by $M$ parts:
$\bm{x} = (\bm{x}^{(1)},\bm{x}^{(2)},\dots,\bm{x}^{(M)} )$, where $\bm{x}^{(m)} \in \mathcal{R}^{d_m}$, $\sum_{m=1}^M d_m = d$, 
and there exist functional relationships between the sub-features $\bm{x}^{(m)}$ and the measured output $y$, that is,  $ y= g_m(\bm{x}^{(m)})+\epsilon^{(m)}$, where $\epsilon^{(m)}$ is the additive noise with zero mean ($E\{\epsilon^{(m)}\} = 0$) and the finite variance ($\textit{Var}\{\epsilon^{(m)}\} = \sigma_{(m)}^2 < \infty$), $m=1,2,\dots, M$.
Thus, we have $y= g(\bm{x}) +\epsilon$, where $g(\bm{x}) = \frac{1}{M} \sum_{m=1}^M g_m( \bm{x}^{(m)} )$, $E\{\epsilon\} = 0$ and $\textit{Var} \{\epsilon \} = \frac{1}{M^2}\sum_{m=1}^M\sigma^2_{(m)} = \sigma^2 $ (under assumption that $\epsilon^{(1)},\epsilon^{(2)},\dots,\epsilon^{(M)}$  are mutually independent random variables). 

Let the $\{f_1, f_2,\cdots,f_M\}$ denote $M$ base models, where the $m$-th model $f_m$ is separately trained on $D_m$, where $D_m =\{(\bm{x}_n^{(m)},y_n), n=1,2,\dots,N\}$.
Usually, the parameters $\theta_m$ of the model $f_m$ is estimated by 
\begin{equation}
\theta_m^*(D_m) = \arg \min_{\theta_m} \frac{1}{N} \sum_{n=1}^{N}(f_m(\bm{x}_n^{(m)}; \theta_m) - y_n)^2.
\end{equation} 
Since the estimated $\theta_m^*$ depends on the given $D_m$, we write $\theta_m^*(D_m)$ to clarify the dependency of $D_m$. 
Therefore, an output value of the $f_m$ for an input $\bm{x}^{(m)}$ should be written as $f_m(\bm{x}^{(m)}, \theta_m^*(D_m))$; 
for simplicity, denoted by $f_m(\bm{x}^{(m)}; D_m)$.
 
The ensemble output for an input $\bm{x}$ is defined as 
\begin{equation}
\label{Eq10_Ensemble_Output}
\bar{f}(\bm{x}) = \sum_{m=1}^{M}a_{m}f_m(\bm{x}^{(m)}; D_m),
\end{equation}
where $\sum_{m=1}^M a_m=1$ and $ 0<a_m<1 $, which is the weight of the $m$-th base model.
For the case of $a_m = 1/M$, the base models independently and equally contribute to the ensemble output, termed as the \textbf{Naive} method in this paper.

Let $(X_0, Y_0)$ be a random sample, taken from $D$ but independent of $D_t$, where $X_0 = [X_0^{(1)}, X_0^{(2)}, \dots, X_0^{(M)}]$.
In the light of the bias/variance decomposition by Geman, Bienenstock and Doursat \cite{geman1992neural}, the ensemble generalization error expression by Ueda and Nakano \cite{ueda1996generalization}, and the diversity discussion by Brown and Wyatt \cite{brown2005managing} for the ensemble modelling,
we extend the result on the generalization error of the ensemble estimator as follows.
\begin{theorem} Let $G_E(\bar{f})$ denote the generalization error of an ensemble $\bar{f}$ in Eq.~(\ref{Eq10_Ensemble_Output}). 
Then, we have
\label{TH02}
\begin{equation}
\label{Eq08_Ensemble_ThreeParts}
G_E(\bar{f}) = E_{X_0}  \left\lbrace \frac{1}{M} \overline{Var}(X_0) + (1 - \frac{1}{M})\overline{Cov}(X_0) + \overline{Bias}(X_0)^2 \right\rbrace + \sigma^2,  
\end{equation}
where $\overline{Var}(X_0)$, $\overline{Cov}(X_0)$ and $\overline{Bias}(X_0)$ are the average conditional variance, conditional covariance and conditional bias of the $M$ models with heterogeneous features for $X_0$, respectively, that is 
\begin{eqnarray}
&& \overline{Var}(X_0)   = \frac{1}{M}\sum_{m=1}^M   Var\left\lbrace (f_m | X_0^{(m)}) \right\rbrace,  \label{Eq09_Var}\\ 
&& \overline{Bias}( X_0) = \frac{1}{M} \sum_{m=1}^M  Bias\left\lbrace (f_m | X_0^{(m)}) \right\rbrace, \\
&& \overline{Cov}(X_0)   = \frac{1}{M(M-1)}\sum_{m=1}^M \sum_{q \neq m } Cov\left\lbrace  (f_m |X_0^{(m)}), (f_q|X_0^{(q)})\right\rbrace. \label{Eq10_Cov}
\label{Eq11_Bias}
\end{eqnarray}
where 
\begin{eqnarray*}
&&  Var\left\lbrace (f_m | X_0^{(m)}) \right\rbrace = E_{D_t} \left\lbrace \left( f_m(X_0^{(m)}; D_t) -E_{D_t}\left\lbrace f_m(X_0^{(m)}; D_t)  \right\rbrace \right)^2 \right\rbrace, \\
&&  Bias\left\lbrace (f_m | X_0^{(m)}) \right\rbrace = E_{D_t} \left\lbrace f_m(X_0^{(m)}; D_t) \right\rbrace  - g_m(X_0^{(m)}), \\
&& Cov\left\lbrace  (f_m, X_0^{(m)}), (f_q | X_0^{(q)}) \right\rbrace = \\
&& E_{D_t} \left\lbrace \left[ f_m (X_0^{(m)}; D_t) - E_{D_t} \left\lbrace f_m (X_0^{(m)}; D_t)\right\rbrace \right] \left[ f_q (X_0^{(q)}; D_t) - E_{D_t} \left\lbrace f_q(X_0^{(q)}; D_t)\right\rbrace \right] \right\rbrace. 
\end{eqnarray*}
\end{theorem}

Therefore, managing the covariance term $\overline{Cov}(X_0)$ explicitly helps in controlling the divergence of the base models, leading to a better generalized ensemble.
In practice, samples from a validation data set $D_v$, which is i.i.d to $D_t$,  could represent the random sample pair $(X_0, Y_0)$ in Theorem 2, and the validation error could be regarded as a realization of the generalization error of the ensemble.

\subsection{Negative correlation learning}
The negative correlation learning (NCL) is a typical training scheme to build neural network ensembles \cite{geman1992neural, hansen1990neural}.
The key idea behind this learning algorithm lies in reducing the covariance among the base models while keeping the variance and bias terms of the ensemble not to be increased.
Mathematically, the cost function of the $m$-th base model over the training data $D_m$ in NCL is given by
\begin{equation}
\label{Eq18_BaseModel_Error}
e_m = \sum_{n=1}^N \frac{1}{2} \left[  (f_m(\bm{x}_n^{(m)})- y_n )^2 + \lambda p_m(\bm{x}_n) \right],
\end{equation}
where
\begin{equation}
\label{Eq13_BaseModel_Penalty}
p_m(\bm{x}_n) = (f_m (\bm{x}_n^{(m)}) - \bar{f}(\bm{x}_n) )\sum_{q\neq m}^M (f_{q}(\bm{x}_n^{(q)}) - \bar{f}(\bm{x}_n)),
\end{equation}
and $0<\lambda<1$ is the regularizing factor. 
Note that
\begin{eqnarray}
\label{Eq14_Ensemble_Output_Average}
\sum_{q \neq m}^M (f_{q}(\bm{x}_n^{(q)})-\bar{f}(\bm{x}_n)) = -(f_{m}(\bm{x}_n^{(m)}) - \bar{f}(\bm{x}_n)).
\end{eqnarray}
Hence, Eq.~(\ref{Eq18_BaseModel_Error}) can be rewritten as
\begin{equation}
\label{Eq15_BaseModel_Error_with_penalty_Simple}
e_m = \sum_{n=1}^N \frac{1}{2}  \left[ (f_m(\bm{x}_n^{(m)})- \bm{y}_n )^2 - \lambda  (f_m (\bm{x}^{(m)}_n) - \bar{f}(\bm{x}_n)^2 \right],  m= 1,2,\dots,M.
\end{equation}
The NCL scheme aims to find all the parameters (including the weights, biases and regularizing factor) of the $M$ models through minimizing every $ e_{m}$.

\section{Stochastic configuration networks ensemble}
In \cite{alhamdoosh2014fast}, RVFL networks are employed as the base models to build a neural network ensemble where all the base models share the same input and identical architecture. 
In the construction of the DNNE, we used the pseudo-inverse method to evaluate the output weights of RVFL models.
However, RVFL networks cannot perform at all if the scope of the random weights and biases  is not chosen properly \cite{li2017insights, WangandLi_SCN, WangandLi_RSCN}. 
Also, it is difficult to estimate the number of hidden node of each RVFL base model in the ensemble.
Due to the merits of the SCN model for overcoming these issues associated with RVFL networks, we employ SCNs as the base learner models with heterogeneous features to build SCNE.

\begin{figure}[t]
\centering
\includegraphics[scale=0.8]{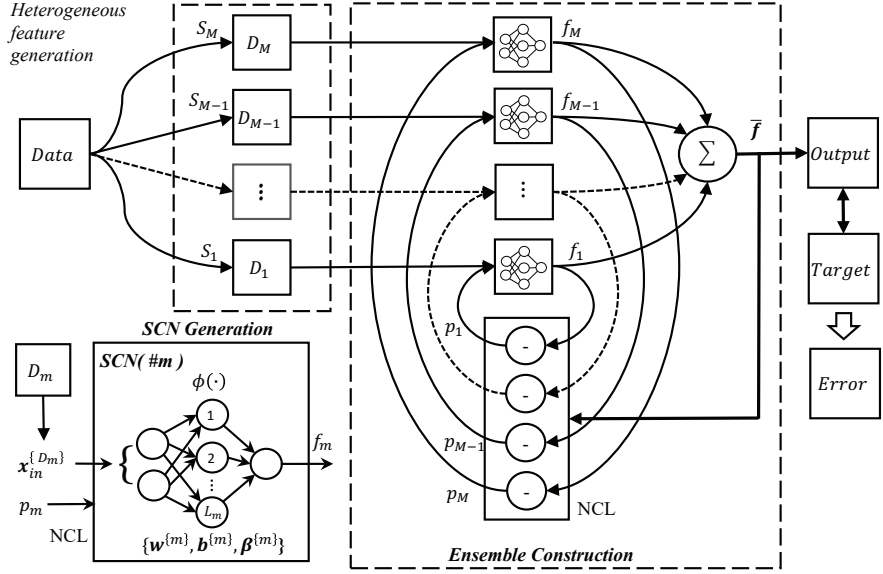}
\caption{Framework of the proposed SCNE with heterogeneous features.}
\label{Fig1_Ensemble}
\end{figure}

For large-scale data modelling, the pseudo-inverse method used in the DNNE for evaluating the output weights of the ensemble is limited. 
Therefore, the \textbf{block Jacobi method} and \textbf{block Gauss-Seidel method} \cite{young1971Iterative}, are used to calculate the output weights of SCNE model. 
The proposed SCNE framework is shown in Fig.~\ref{Fig1_Ensemble}. 
The original data set $D$ is partitioned into $M$ subsets according to the heterogeneous feature sets, denoted by $\{S_m, m=1,2,\cdots,M\}$.  
Then, the corresponding sample data set for the $m$-th model is denoted as $D_m = \{ (\bm{x}_n^{(m)}, y_n), n=1,2,\cdots,N\}$, where $\bm{x}_n^{(m)} \in \mathcal{R}^{d_m}$ and $y_n\in \mathcal{R}$. 
Each set will be used to build the SCN base model in the ensemble.
The partition of the full feature set could be done by a prior knowledge or various feature selection schemes. 

The procedure of building the SCNE model could be divided into two steps:
(i) the SCN base model generation, that is, each SCN model will be built independently according to the SC-III algorithm in \cite{WangandLi_SCN}; and (ii) the ensemble construction by means of the NCL algorithm, where the random weights and biases of the SCN models generated in the first step are fixed.

\subsection{SCN base model generation}
The SCN base models can be constructed individually with $L_m$ nodes in the hidden layer on the data set $D_m$, and the output weights $\bm{\beta}_m$ are also calculated according the stochastic configuration algorithm (SC-III) in \cite{WangandLi_SCN}.
One of the important issues in building SCNs is associated with the overfitting problem when too many hidden nodes are added incrementally.
Fortunately, this problem can be solved by monitoring the modelling performance over a validation set with a quick pruning method. 
Alternatively, we can apply the trial-and-error method to determine an appropriate number of the hidden nodes of the SCN base model.

\subsection{SCNE construction}
In the ensemble construction, the input weights and biases of the base models are fixed. 
The output weights $\bm{\beta}_1, \bm{\beta}_2, \dots, \bm{\beta}_M$ of the SCNE model can be obtained by using the NCL scheme, that is, 
\begin{equation}
\label{Eq16_TheB}
\{ \bm{\beta}_1^*, \bm{\beta}_2^* \dots, \bm{\beta}_M^* \} =\arg \min_{\bm{\beta}_m^*}\{e_m\}, m= 1,2,\dots,M.
\end{equation}
The following sections detail the pseudo-inverse method and two iterative methods for computing the output weights.
To understand some properties of the solutions obtained by these three methods, a demonstration over a 2-D function approximation is given to see the time cost and the impact of the regularized model on the correlationship among the solutions.

\subsubsection{Analytical solution}
By Eq.~(\ref{Eq15_BaseModel_Error_with_penalty_Simple}), the cost function $e_m$ of the SCN model $f_m$ with the NCL penalty term could be rewritten in the following matrix form:
\begin{equation} 
\label{Eq17_em}
e_m = \frac{1}{2} \left( \Vert H_m \bm{\beta}_m - \bm{y} \Vert^2 - \lambda \Vert H_m \bm{\beta}_m - \frac{1}{M} \bm{H} \bm{B} \Vert^2 \right),
\end{equation}
where 
\begin{equation} \label{Eq18_Hm}
H_m =
\left[
\begin{array}{ccc}
\phi_1(\bm{x}_1^{(m)}) & \ldots & \phi_{L_m}(\bm{x}_1^{(m)}) \\
\vdots & \ddots & \vdots \\
\phi_{1}(\bm{x}_N^{(m)})  & \ldots & \phi_{L_m}(\bm{x}_N^{(m)}) 
\end{array} \right]_{N \times L_m} ,
\bm{B} = \left[
\begin{array}{c}
\bm{\beta}_1  \\
\vdots  \\
\bm{\beta}_M
\end{array} \right] _{\bm{L} \times 1}, 
\bm{y} =  
\left[
\begin{array}{c}
y_1  \\
\vdots \\
y_N
\end{array}
\right]_{N \times 1}.
\end{equation}
The $H_m$ is the output matrix at the hidden layer of the $m$-th base model; $\bm{B}$ is the output weighs of the whole ensemble (all the output weights of the base models); $\bm{y}$ is the target; $\bm{H} = \left[H_1 , H_2, \dots , H_M \right]_{N \times \bm{L}}$; $\bm{L}$ is the total number of the hidden nodes the SCNE model.
Simple computations give that
\begin{equation}
\label{Eq22_em_derivative}
\frac{\partial {e_m}}{\partial{\beta_m}} = 
c_1
H_m^T H_m \bm{\beta}_m
+
c_2
H_m^T\bm{\tilde{H}}_m\bm{B} 
-
H_m^T\bm{y}=0,\quad m = 1,2,\dots,M,
\end{equation}
where $\bm{\tilde{H}}_m  = \left[ H_1, \ldots,H_{m-1}, \bm{0}_{\tiny N \times L_m}, H_{m+1}, \ldots, H_M \right]_{N \times \bm{L}}$, and 
\begin{eqnarray}
&& c_1 = 1-\frac{\lambda(M-1)^2}{M^2} ,\quad
c_2 = \frac{\lambda(M-1)}{M^2}. \label{Eq23_c1c2}
\end{eqnarray}
From Eq.~(\ref{Eq22_em_derivative}), a huge sized linear equation system can be obtained, that is,
\begin{equation}\label{Eq25_HB_HY}
\mathbb{H}\bm{B} = \bm{H}^T\bm{y}, 
\end{equation}
where 
\begin{equation}\label{Eq26_HHLL}
\mathbb{H} = 
\left[
\begin{array}{cccc}
c_1H_1^TH_1 & c_2H_1^TH_2 & \ldots & c_2H_1^TH_M \\
c_2H_2^TH_1 & c_1H_2^TH_2 & \ldots & c_2H_2^TH_M \\
\vdots & \vdots & \ddots & \vdots \\
c_2H_M^TH_1 & c_2H_M^TH_2 & \ldots & c_1H_M^TH_M
\end{array}
\right]_{\bm{L} \times \bm{L}}.
\end{equation}
If all the $H_m^TH_m$s are invertible, which means all the SCN models are built with a full column rank of $H_m$, then $\mathbb{H}$ is invertible if the regularizing factor $\lambda$ takes a sufficiently smaller value. In this case, $\bm{B}=\mathbb{H}^{-1}\bm{H}^T\bm{y}$.
However, we cannot ensure full rank condition for every matrix $H_m^TH_m$ in system implementation. 
In such a case, the output weights of the SCNE model can be evaluated by
\begin{equation}
\label{Eq27_B_HHY}
\bm{B}=\mathbb{H}^{\dagger}\bm{H}^T\bm{y},
\end{equation}
where the $\mathbb{H}^{\dagger}$ represents the pseudo-inverse of the matrix $\mathbb{H}$ \cite{golub1996Matrix}.\\

{\centering
\begin{minipage}{0.8\linewidth}
\begin{algorithm}[H]
\SetAlgoLined
\SetKwInOut{Input}{Input}\SetKwInOut{Output}{Output}
\caption{Pseudo-inverse method for building SCNE}
\label{Algorithm 1}
\DontPrintSemicolon
\Input{$M$, $\{L_m\}$, $ \{D_m\} = \{(\bm{x}_n^{(m)}, \bm{y}), n = 1,2, \dots, N\}$, $\bm{x}_n^{(m)} \in \mathcal{R}^{d_m}$, $\bm{y}\in \mathcal{R}$; $\lambda$;} 
\Output{SCNE.}
\BlankLine
\textbf{SCN generation by SC-III}\;
\textbf{Calculate all $\{H_m\}$ and $\bm{H}$} (Eq.~(\ref{Eq18_Hm}))\;
\
\For{$m \leftarrow 1$ to $M$ }{
     \For{$n \leftarrow 1$ to $N$, $l \leftarrow 1$ to $L_m$}{
           $H_m(n,l) = \phi_l(\bm{x}_n^{(m)})$ 
        }
    $\bm{H} := [\bm{H},  H_m]$}
\textbf{Calculate the coefficients $c_1$ and $c_2$} (Eq.~(\ref{Eq23_c1c2}))\; 
$c_1 = 1- \lambda(M-1)^2/M^2$,\quad $c_2 = \lambda(M-1)/M^2$\;
\textbf{Calculate the large matrix $\mathbb{H}$} (Eq.~(\ref{Eq26_HHLL}))\;
 \For{ $m,q \leftarrow 1$ to $M$ }{
     \eIf{$m \neq q$}{$\mathbb{H}(m,q) = c_2 H_m^T H_q$}
     {$\mathbb{H}(m,q) = c_1 H_m^T H_q$}
    }
\textbf{return} $\bm{B}=\mathbb{H}^{\dagger}\bm{H}^T\bm{y}$ (Eq.~(\ref{Eq27_B_HHY})).
\end{algorithm}
\end{minipage}
\par
}\vspace{5mm}

\subsubsection{Iterative solutions}
Due to some constraints on either the numerical algorithms for computing the pseudo inverse of a huge sized matrix or computing device with limited RAM memory, the analytical solution described in \textbf{Algorithm~\ref{Algorithm 1}} becomes impractical and/or time-consuming for large-scale datasets.  Therefore, it is necessary and important to develop iterative algorithms for problem solving. Instead of evaluating the whole output weights $\bm{B}$ at once, we update gradually the $\bm{\beta}$ of each SCN base model to reduce the computing time and memory cost.  

The minimum of the cost function Eq. (\ref{Eq17_em}) with respect to the local output weights $\bm{\beta_m}$  at $k$-th iteration  can be obtained by solving the following equation: 
\begin{equation}
\label{Eq28_Beta_mp}
\frac{\partial e_{m}^{(k)}}{\partial  \bm{\beta}_m}=  
\frac{\partial \left( \Vert H_m \bm{\beta}_m - \bm{y} \Vert^2 - \lambda \Vert H_m \bm{\beta}_m - \frac{1}{M} \bm{H} \bm{B}_{m,X}^{(k)} \Vert ^2 \right) }{ 2 \partial \bm{\beta}_m} = 0,
\end{equation} 
where $X\in \{J,G\}$, indexing the \textbf{Jacobi} and  \textbf{Gauss-Seidel} iteration schemes, respectively \cite{young1971Iterative}.

Taking the initial values of the output weights as $\bm{B}^{(0)} = \left[H_1^{\dagger}\bm{y},H_2^{\dagger}\bm{y},\cdots, H_M^{\dagger}\bm{y} \right]^T$. From Eq. (\ref{Eq28_Beta_mp}), we can derive the following block iterative algorithms:
\begin{eqnarray}
&&\bm{\beta}_{m,J}^{(k)} = \frac{1}{c_1}H_m^{\dagger} \left( \bm{y} - c_2 \bm{\tilde{H}}_m\bm{B}_{m,J}^{(k)}\right), \label{Eq30_Synchronous_method}
\\  
&&\bm{B}_{m,J}^{(k)} = \left[
\bm{\beta}_1^{(k-1)},  
\cdots,
\bm{\beta}_{m-1}^{(k-1)},
\bm{\beta}_{m},
\bm{\beta}_{m+1}^{(k-1)},
\cdots,
\bm{\beta}_M^{(k-1)}
\right]^T, \\
&&\bm{\beta}_{m,G}^{(k)} = \frac{1}{c_1}H_m^{\dagger} \left( \bm{y} - c_2 \bm{\tilde{H}}_m\bm{B}_{m,G}^{(k)}\right), \label{Eq31_Asynchronous_method}
\\
&&\bm{B}_{m,G}^{(k)} = \left[
\bm{\beta}_1^{(k)},
\cdots,
\bm{\beta}_{m-1}^{(k)},
\bm{\beta}_{m},
\bm{\beta}_{m+1}^{(k-1)},
\cdots,
\bm{\beta}_M^{(k-1)}
\right]^T,
\end{eqnarray}

The pseudo codes of these two algorithms are given in \textbf{Algorithms~\ref{Algorithm 2}} and \textbf{\ref{Algorithm 3}} below. 
The main difference between these two schemes occurs at line 13. In practice, we provide a \textbf{tolerance $\tau$} and a \textbf{ maximum iteration number $k_{max}$ } experimentally to guarantee that the algorithms will be terminated if either the SCNE error $E_{ens} < \tau$ or the iteration time reaches the $k_{max}$. Note that Eqs.~(\ref{Eq30_Synchronous_method}) and (\ref{Eq31_Asynchronous_method}) are rewritten in discrete summation forms in the algorithms for the convenience of programming.

{\centering
\begin{minipage}{0.8\linewidth}
\begin{algorithm}[H]
\SetAlgoLined
\SetKwInOut{Input}{Input}\SetKwInOut{Output}{Output}
\caption{Block Jacobi method for building SCNE}
\label{Algorithm 2}
\DontPrintSemicolon
\Input{$M$, $\{L_m\}$, $ \{D_m\} = \{(\bm{x}_n^{(m)}, \bm{y}), n = 1,2, \dots, N\}$, $\bm{x}_n^{(m)} \in \mathcal{R}^{d_m}$, $\bm{y}\in \mathcal{R}$; $\tau$; $k_{max}$;} 
\Output{ SCNE.}
\BlankLine
\textbf{SCN generation by SC-III}\;
\textbf{Calculate all $\{H_m\}$} and $\{\bm{\beta}_m^{(0)}\}$\;
\For{$m \leftarrow 1$ to $M$ }{
     \For{$n \leftarrow 1$ to $N$, $l \leftarrow 1$ to $L_m$}{
           $H_m(n,l) = \phi_l(\bm{x}_n^{(m)})$ 
        }
     $\bm{\beta}_m^{(0)} = H_m^{\dagger}\bm{y}$ \;   
    }
\textbf{Calculate the coefficients $c_1$ and $c_2$}\; 
$c_1 = 1- \lambda(M-1)^2/M^2$,\quad $c_2 = \lambda(M-1)/M^2$\;
\For {$k \leftarrow 1$ to $k_{max}$}{
    \For{$m \leftarrow 1$ to $M$ }{ $\bm{\beta}_{m,J}^{(k)} = \frac{1}{c_1}H_m^{\dagger} \left( \bm{y} - c_2 \sum_{q \neq m}{H_{q}\bm{\beta}_{q}^{(k-1)}}\right)$  (Eq.~(\ref{Eq30_Synchronous_method}))\;
    Update $\bm{B}_J^{k}$ with $\bm{\beta}_{m,J}^{k}$
    }
    \If { $E_{ens}^{(k)}  < \tau$}{break}
}
\textbf{return} $\bm{B}_J \leftarrow \bm{B}_J^{k}$. 
\end{algorithm}
\end{minipage}
\par
}

\begin{remark}
Suppose that every $H_m$ has full column rank. Then, under some certain conditions, we can prove that  the block Jacobi and block Gauss-Seidel methods  \cite{young1971Iterative} converge to the same solution as obtained by the pseudo-inverse method. To do so, let us rewrite  the block Jacobi iterative scheme in the the following matrix form:
\begin{equation}\label{Eq32_Bk_DR}
 \bm{B}_J^{(k)} = \mathbb{D}^{-1} (\bm{H}^T\bm{y} - \mathbb{R} \bm{B}_J^{(k-1)}), 
\end{equation}
where
\begin{equation}\label{Eq33_D}
\mathbb{D} = 
\left[
\begin{array}{cccc}
c_1H_1^TH_1 & 0 & \ldots & 0 \\
0 & c_1H_2^TH_2 & \ldots & 0\\
\vdots & \vdots & \ddots & \vdots \\
0 & 0 & \ldots & c_1H_M^TH_M
\end{array}
\right],
\mathbb{R} = 
\left[
\begin{array}{cccc}
0 & c_2H_1^TH_2 & \ldots & c_2H_1^TH_M \\
c_2H_2^TH_1 & 0 & \ldots & c_2H_2^TH_M \\
\vdots & \vdots & \ddots & \vdots \\
c_2H_M^TH_1 & c_2H_M^TH_2 & \ldots & 0
\end{array}
\right].
\end{equation}
Similarly,  the block Gauss-Seidel iterative scheme can be rewritten as 
\begin{equation}\label{Eq35_Bk_LU}
\bm{B}_G^{(k)} = \mathbb{L}^{-1} (\bm{H}^T\bm{y} - \mathbb{U} \bm{B}_G^{(k-1)}), 
\end{equation}
where
\begin{equation} \label{Eq36_L}
\mathbb{L} = 
\left[
\begin{array}{cccc}
c_1H_1^TH_1 & 0 & \ldots & 0 \\
c_2H_2^TH_1 & c_1H_2^TH_2 & \ldots & 0\\
\vdots & \vdots & \ddots & \vdots \\
c_2H_M^TH_1 & c_2H_M^TH_2 & \ldots & c_1H_M^TH_M
\end{array}
\right],
\mathbb{U} = 
\left[
\begin{array}{cccc}
0 & c_2H_1^TH_2 & \ldots & c_2H_1^TH_M \\
0 & 0 & \ldots & c_2H_2^TH_M \\
\vdots & \vdots & \ddots & \vdots \\
0 & 0 & \ldots & 0
\end{array}
\right].
\end{equation}
Suppose that $\bm{B}_{J}^* := \lim_{k\rightarrow \infty}\bm{B}_{J}^{(k)}$ and $\bm{B}_{G}^* := \lim_{k\rightarrow \infty}\bm{B}_{G}^{(k)}$ exist. Then, from Eq.~(\ref{Eq32_Bk_DR}), the following holds
\begin{equation}
\label{Eq38_limBs}
\lim_{k\rightarrow \infty} \bm{B}_J^{(k)} = \lim_{k\rightarrow \infty} \mathbb{D}^{-1}(\bm{H}^T\bm{y} - \mathbb{R} \bm{B}_J^{(k-1)}).
\end{equation}
Therefore, $(\mathbb{D} + \mathbb{R})\bm{B}_J^* = \bm{H}^T\bm{y}$.
Similarly, we have $(\mathbb{L} + \mathbb{U})\bm{B}_G^* = \bm{H}^T\bm{y}$.

Note that $\mathbb{H} = \mathbb{D} + \mathbb{R} = \mathbb{L} + \mathbb{U}$, where $\mathbb{L}$ is lower triangular component of $\mathbb{H}$,  
$\mathbb{U}$ is the strictly upper triangular component, $\mathbb{D}$ is a diagonal component of $\mathbb{H}$, and $\mathbb{R}$ is the remainder. 
Thus, we have
\begin{equation}
\label{Eq40_BsBaB}
(\mathbb{D} + \mathbb{R})\bm{B}_J^* = (\mathbb{L} + \mathbb{U})\bm{B}_G^* = \mathbb{H}\bm{B} = \bm{H}^T\bm{y}.
\end{equation}
From Eq.~(\ref{Eq40_BsBaB}), we get $\mathbb{H}(\bm{B}_J^*-\bm{B}) = \mathbb{H}(\bm{B}_G^*- \bm{B}) = \bm{0}$.
If $\mathbb{H}$ is nonsingular, it is easy to see that $\bm{B}_J^* = \bm{B}_G^* = \bm{B}$. Thus, we conclude that the two iterative schemes converge to the same solution from the pseudo-inverse method provided that (i) every $H_m$ is full column rank; (ii) the regularizing factor $\lambda$ is chosen so that 
 the spectral radius of the iteration matrix is less than 1, that is,
\begin{equation}
\label{Eq41_Convergence}
\rho(\mathbb{D}^{-1}\mathbb{R}) < 1 \quad or \quad \rho(\mathbb{L}^{-1}\mathbb{U}) < 1.
\end{equation}
By using Greschgorin Circle Theorem \cite{golub1996Matrix} for block matrices, we know that any eigenvalue $z$ of the matrix $\mathbb{D}^{-1}\mathbb{R}$ must be constrained by one of the following inequalities:
\begin{equation}
|z| \leq \frac{c_2}{c_1}\sum_{q \ne m} \norm{(H_m^TH_m)^{-1} H_m^TH_q},  m=1,2,\dots, M.
\end{equation}
Hence, a theoretical estimate of the regularizing factor $\lambda$ should be subjected to $c_2c_1^{-1}\theta_0<1$, where 
\begin{equation}
\theta_0 = \max \{\sum_{q \ne m} \norm{(H_m^TH_m)^{-1} H_m^TH_q},   q=1,2,\dots, M \}.
\end{equation}
Simple computations give that
\begin{equation}
0<\lambda<\frac{M^2}{(M-1)(M+\theta_0-1)}.
\end{equation}
\end{remark}

{\centering
\begin{minipage}{0.8\linewidth}
\begin{algorithm}[H]
\SetAlgoLined
\SetKwInOut{Input}{Input}\SetKwInOut{Output}{Output}
\caption{Block Gauss-Seidel method for building SCNE}
\label{Algorithm 3}
\DontPrintSemicolon
\Input{$M$, $\{L_m\}$, $ \{D_m\} = \{(\bm{x}_n^{(m)},\bm{y}), n = 1,2, \dots, N\}$, $\bm{x}_n^{(m)} \in \mathcal{R}^{d_m}$, $\bm{y}\in \mathcal{R}$; $\phi$; $\alpha$; $\lambda$; $\tau$; $k_{max}$;} 
\Output{ SCNE.}
\BlankLine
\textbf{SCN generation by SC-III}\;
\textbf{Calculate all $\{H_m\}$} and $\{\bm{\beta}_m^{(0)}\}$\;
\For{$m \leftarrow 1$ to $M$ }{
     \For{$n \leftarrow 1$ to $N$, $l \leftarrow 1$ to $L_m$}{
           $H_m(n,l) \leftarrow \phi_l(\bm{x}_n^{(m)})$ 
        }
     $\bm{\beta}_m^{(0)} = H_m^{\dagger}\bm{y}$ \;   
    }
\textbf{Calculate the coefficients $c_1$ and $c_2$}\; 
$c_1 = 1- \lambda(M-1)^2/M^2$,\quad $c_2 = \lambda(M-1)/M^2$\;
\For {$k \leftarrow 1$ to $k_{max}$}{
    \For{$m \leftarrow 1$ to $M$ }{  $\bm{\beta}_{m,G}^{(k)} =\frac{1}{c_1}H_m^{\dagger} \left[ \bm{y} - c_2 \left( \sum_{q < m}{H_{q}\bm{\beta}_{q}^{(k)}} + \sum_{q > m}{H_{q}\bm{\beta}_{q}^{(k-1)}} \right)\right]$  (Eq.~(\ref{Eq31_Asynchronous_method}))\;            
    Update $\bm{B}_G^{k}$ with $\bm{\beta}_{m,G}^{k}$            
    }
    \If {$E_{ens}^{(k)} < \tau$}{ break}
}
\textbf{return} $\bm{B}_G \leftarrow \bm{B}_G^{k}$.
\end{algorithm}
\end{minipage}
\par
}

\begin{remark}
In practice, the full rank condition is hard to meet for all the base models. Thus, the convergence and  uniqueness of the iterative solutions cannot be ensured. To fix this problem, we can employ a regularization model to resolve the linear model in Eq.~(\ref{Eq25_HB_HY}), resulting in a solution as $\bm{B}=\mathbb{H}_r^{-1}\bm{H}^T\bm{y}$, where
\begin{equation}
\label{Eq43_HR}
\mathbb{H}_r = 
\left[
\begin{array}{cccc}
c_1H_1^TH_1 & c_2H_1^TH_2 & \ldots & c_2H_1^TH_M \\
c_2H_2^TH_1 & c_1H_2^TH_2 & \ldots & c_2H_2^TH_M \\
\vdots & \vdots & \ddots & \vdots \\
c_2H_M^TH_1 & c_2H_M^TH_2 & \ldots & c_1H_M^TH_M
\end{array}
\right]
+c_1  
\left[
\begin{array}{cccc}
r_1I_1 & 0 & \ldots & 0\\
0 & r_2I_2 & \ldots & 0\\
\vdots & \vdots & \ddots & \vdots\\
0 & 0 & \ldots & r_M I_M\\
\end{array}
\right],
\end{equation}
here $r_m$ is a positive regularization parameter of the $m$-th SCN base model, and $I_m$ is a $L_m \times L_m$ identity matrix, $m = 1, 2, \dots, M$.
Obviously, the  $\mathbb{H}_r$ is invertible. 
For the iterative methods, each $H_m^{\dagger}$ is replaced by $(H_m^TH_m + r_m I_m)^{-1}H_m^T$ in Eqs.~(\ref{Eq30_Synchronous_method}) and (\ref{Eq31_Asynchronous_method}),  the new initial $\bm{B^{(0)}}$ and the iterative equations of $\bm{\beta}_{m}$ can be updated as follows:
\begin{eqnarray}
\bm{B}^{(0)} = \left[
\begin{array}{c}
\bm{\beta}_1^{(0)}  \\
\vdots  \\
\bm{\beta}_M^{(0)} 
\end{array} \right] 
= \left[
\begin{array}{c}
(H_1^TH_1 + r_1 I_1)^{-1}H_1^T \bm{y}  \\
\vdots  \\
(H_M^TH_M + r_M I_M)^{-1}H_M^T \bm{y} 
\end{array} \right], \label{Eq45_B0}   \\
\bm{\beta}_{m,J}^{(k)} = \frac{1}{c_1}(H_m^TH_m + r_m I_m)^{-1}H_m^T \left( \bm{y} - c_2 \bm{\tilde{H}}_m\bm{B}_{m,J}^{(k)}\right), \label{Eq46_BMS}  \\
\bm{\beta}_{m,G}^{(k)} = \frac{1}{c_1}(H_m^TH_m + r_m I_m)^{-1}H_m^T \left( \bm{y} - c_2 \bm{\tilde{H}}_m\bm{B}_{m,G}^{(k)}\right).\label{Eq47_BMA}  
\end{eqnarray}
To reduce the algorithm complexity, we take equal values for $r_1, r_2,\dots, r_M$, namely $r$, in Eq. (\ref{Eq43_HR}). Adjusting the $r$ will control the $l_2$-norm of the output weights of the ensemble model, especially when dealing a large number of hidden nodes. It is interesting to look into the impact of such a parameter on the modelling performance. It is still unclear on this point, and a further research is being expected in this direction.
\end{remark}

\subsection{Demonstration}
\label{CompareAlgorithms}
A  synthetic dataset is used to demonstrate  the training time along with the number of the hidden nodes $\bm{L}$ in constructing the SCNEs with the pseudo-inverse, block Jacobi and block Gauss-Seidel methods, respectively. The reason behind  this set-up is that a normal PC can run this task as the $\bm{L}$ grows up, and the memory cost of the computer is controllable compared to the large datasets. In this section, we report two sets of results obtained from the original SCNE and the modified SCNE with a regularization factor $r=0.1$. From this comparison, we can infer that a similar consequence holds for 
 large-scale datasets. 
\begin{figure}[h!]
\centering
    \begin{subfigure}[b]{0.49\textwidth}
        \includegraphics[width=\textwidth]{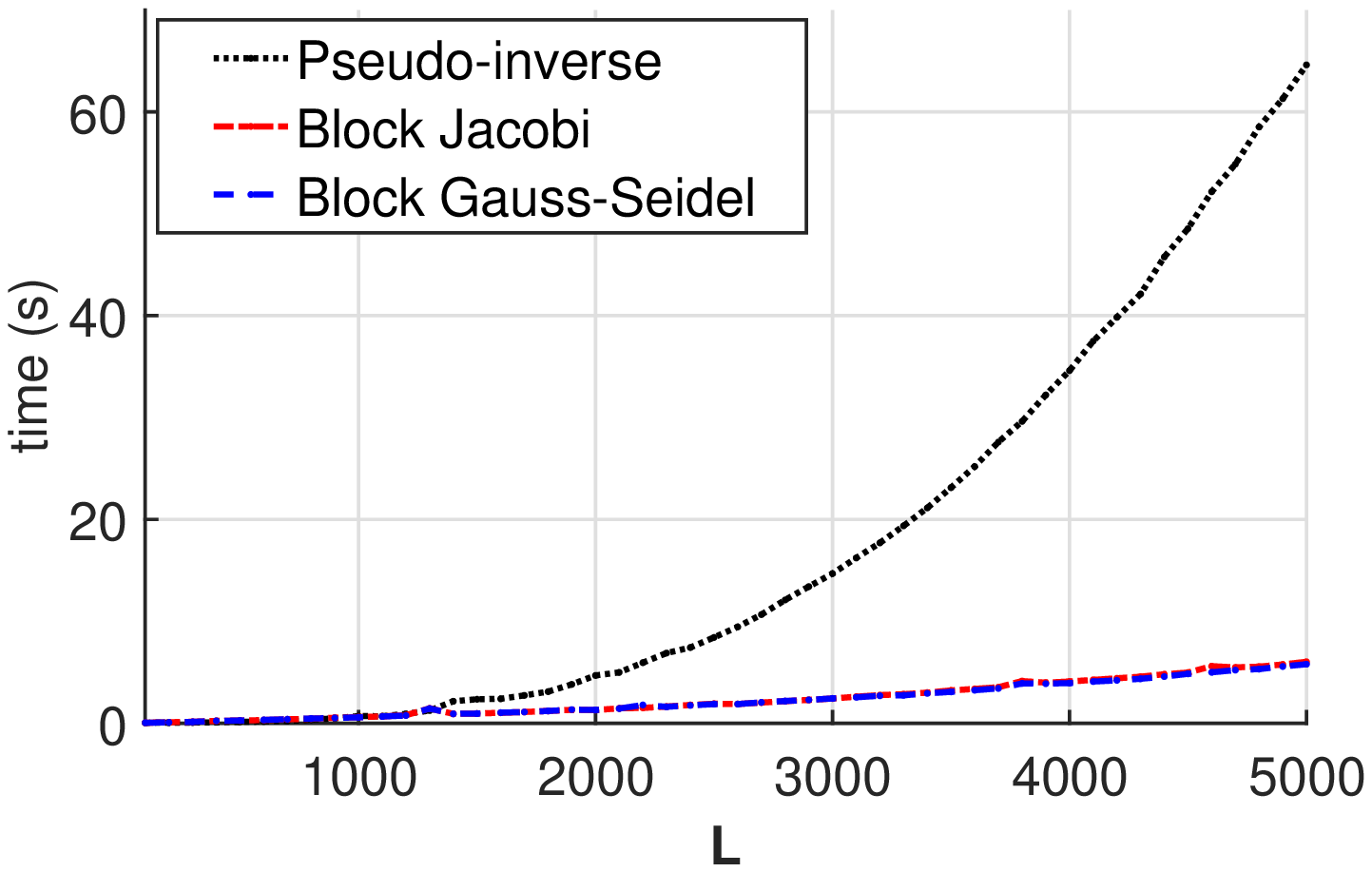}
        \caption{ }
        \label{Fig04a}    
    \end{subfigure}
    \begin{subfigure}[b]{0.49\textwidth}
        \includegraphics[width=\textwidth]{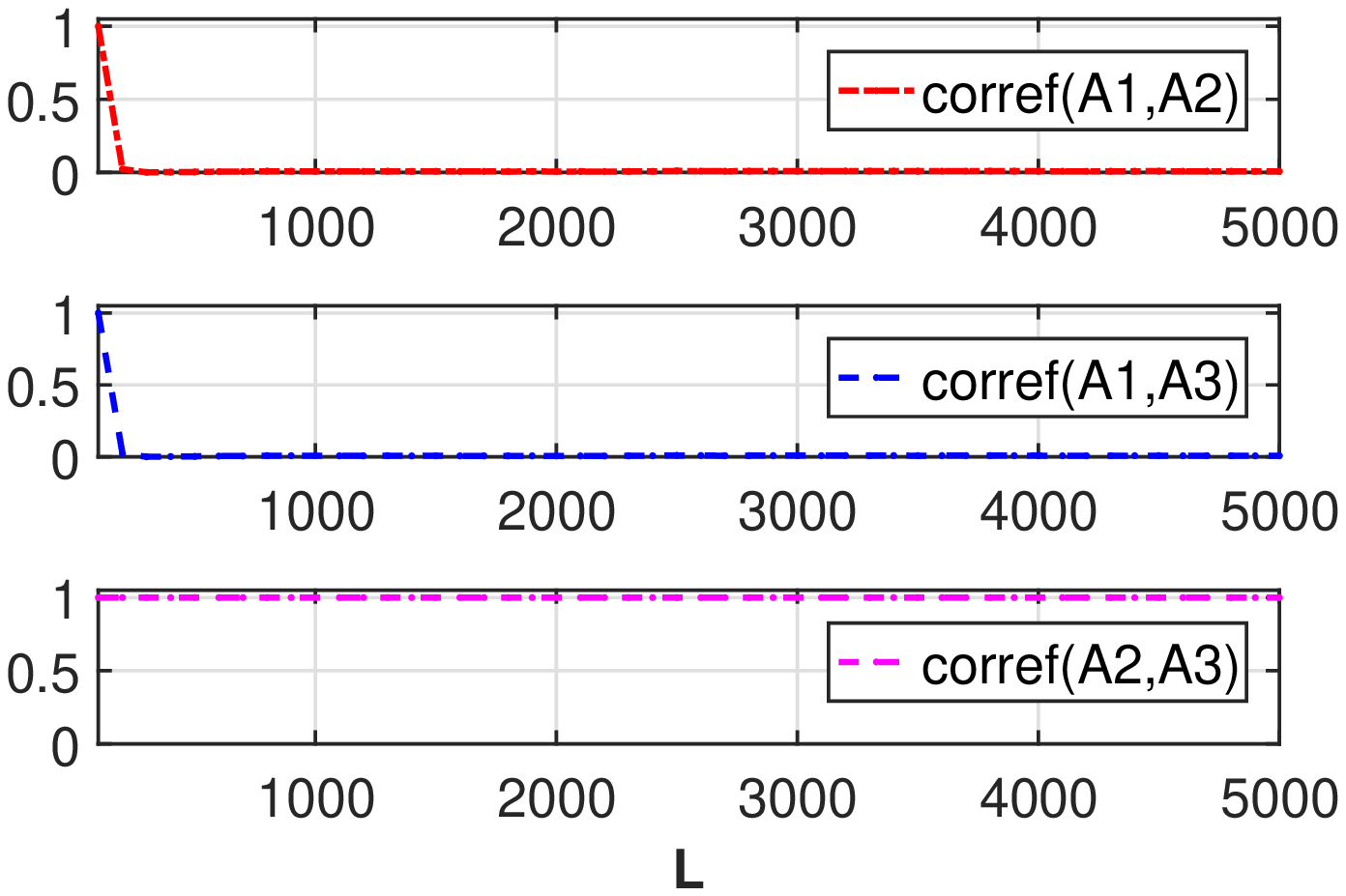}
        \caption{ }
        \label{Fig04b}    
    \end{subfigure}        
    \begin{subfigure}[b]{0.49\textwidth}
        \includegraphics[width=\textwidth]{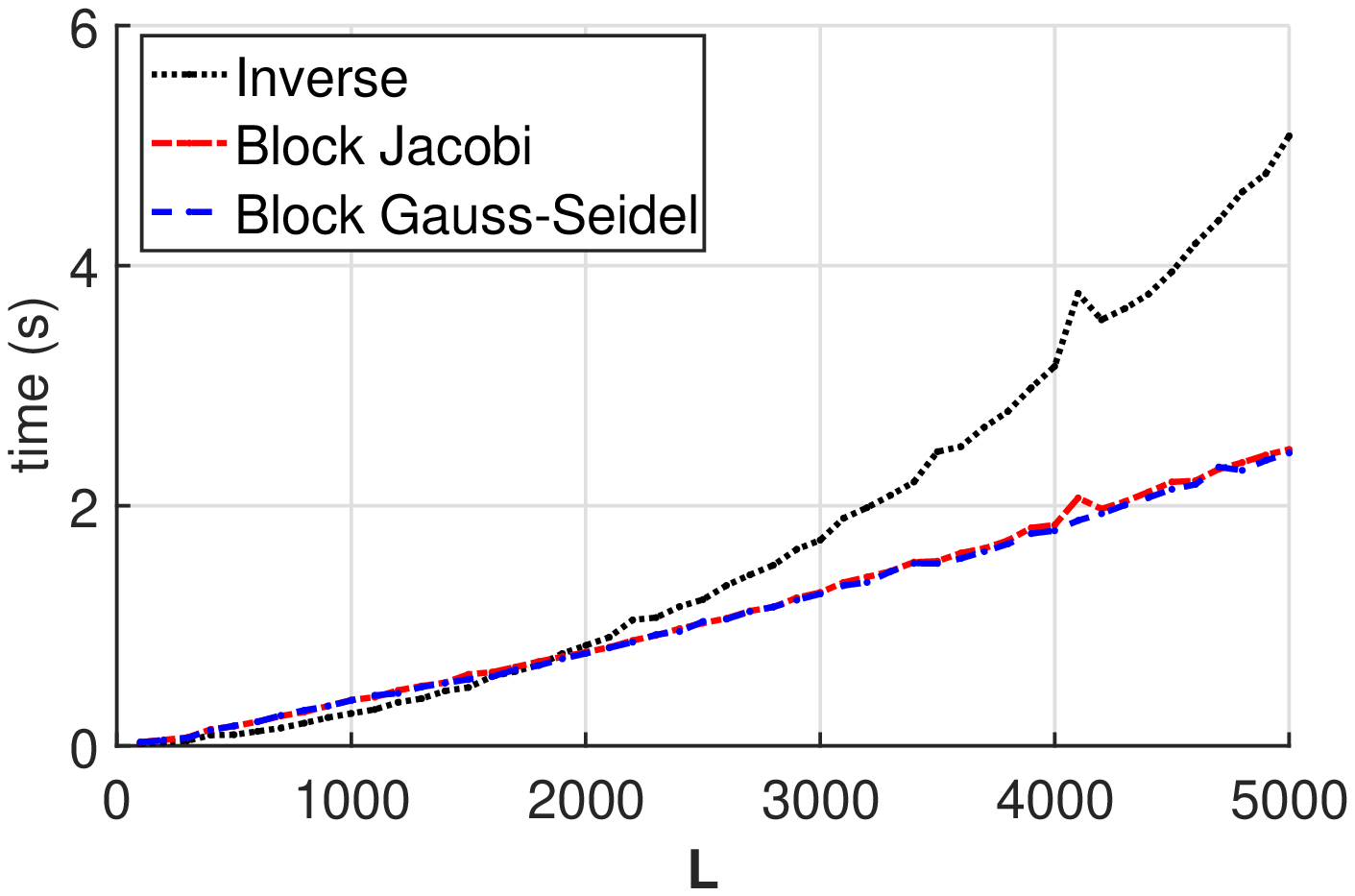}
        \caption{ }
        \label{Fig04c}    
    \end{subfigure}
    \begin{subfigure}[b]{0.49\textwidth}
        \includegraphics[width=\textwidth]{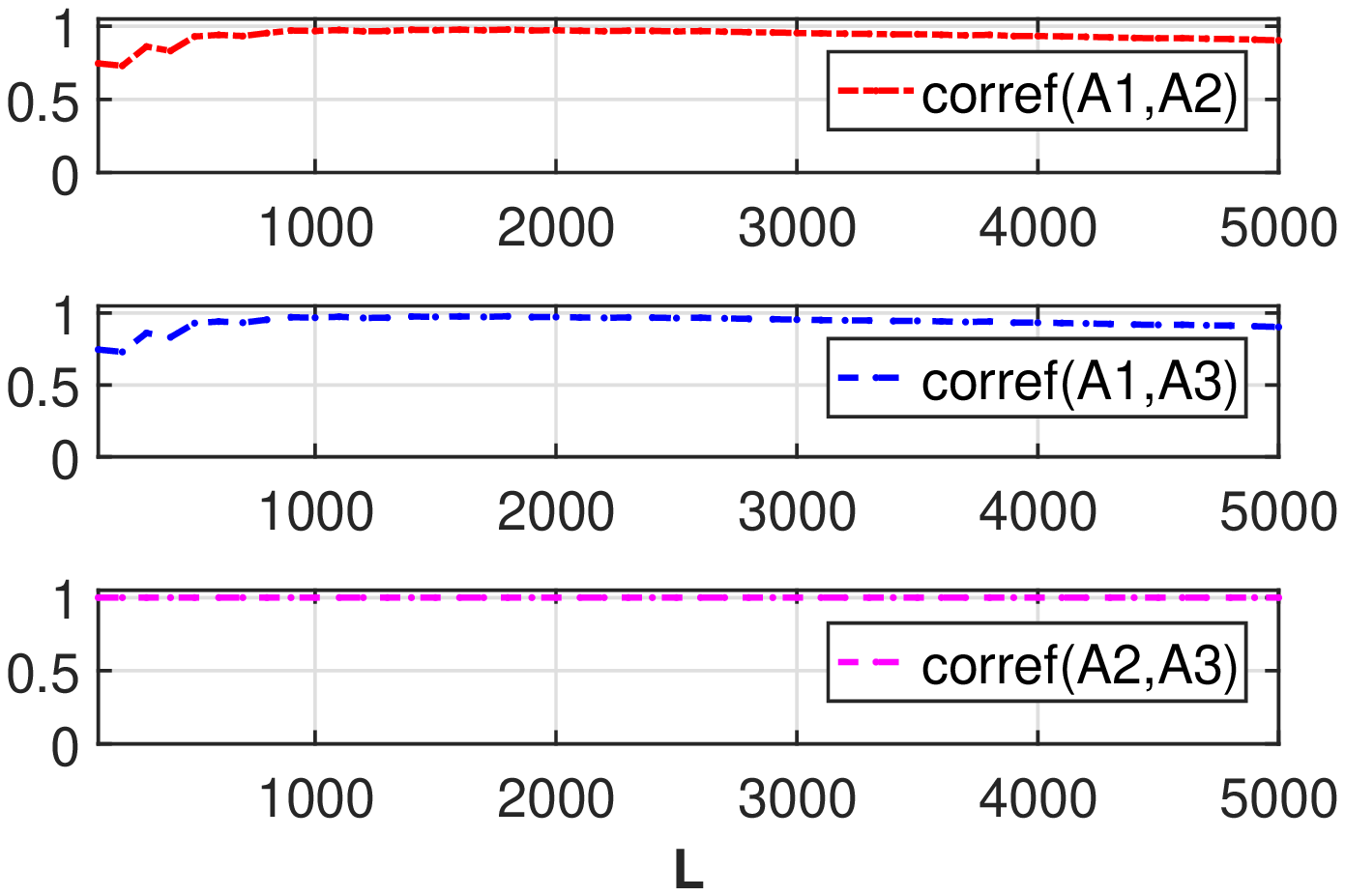}
        \caption{ }
        \label{Fig04d}    
    \end{subfigure}               
\caption{ Comparisons of different algorithms (a) Ensemble construction time by Algorithm 1-3; (b) Correlation coefficients of $\mathbf{B}$s from Algorithm 1-3 (c)  Ensemble construction time of Algorithm 1-3 with regularization ($r = 0.1$) (d) Correlation coefficients of the $\mathbf{B}$s from Algorithm 1-3 with regularization.}
\label{Fig04:CompareAlgorithms}
\end{figure}

The synthetic dataset $D = \{ (x_{1n},x_{2n}, y_n), n =1,...,5000\}$ contains 4,000 training samples and 1,000 test samples, all generated by a simple function $y=cos(2\times x_2)/e^{x_1}$ where $x_2 = sin(x_1)$, $x_1 \in [-5,5]$.  
This dataset is easy to learn by the SCNE, so we omit the training and test results in the context in order to focus on the time comparisons.
In this demonstration, all the SCN base models use the same training datasets.
By setting the $M=10$ (10 base models), $\lambda = 0.1$, $\tau = 10^{-6}$, $k_{max} = 5$, and gradually increasing the hidden node of each base model $L_m$ from $10$ to $500$, the computing time spent on SCNEs construction by using the three algorithms is presented in Fig.~\ref{Fig04a} (original) and Fig.~\ref{Fig04c} (regularized), respectively.

It is clear to see that the computing time of the pseudo-inverse method grows exponentially, while the computing time of the iterative methods grow linearly. The iterative methods may cost more time when the hidden node is small, but when $\bm{L}$ becomes larger ($\bm{L} > 2000$ in this demonstration), the pseudo-inverse approach is clearly inefficient and infeasible for large-scale datasets.
 
With regard to the convergence of the output weights $\bm{B}$ from different algorithms, Fig.~\ref{Fig04b} shows the correlations between these three algorithms. Here, A1, A2 and A3 represent the pseudo-inverse,  the block Jacobi  and  the block Gauss-Seidel methods, respectively. 
The results show that with these settings, the correlations of the A1-A2 and A1-A3 decrease along with the increasing $\bm{L}$, meanwhile, the correlation coefficient of A2-A3 is  consistent, implying that the iterative methods  converge into the same solution.
This phenomenon is caused by the use of the pseudo-inverse in evaluating the output weights.
With the increasing $\bm{L}$, the rank of $\mathbb{H}$ drops sharply, leading to an uncorrelated result to the iterative methods.  
When all the pseudo-inverse calculations of the algorithms are replaced by a regularized form with $r = 0.1$, the correlations of $\bm{B}$ between all these algorithms are almost one, as  shown in Fig.~\ref{Fig04d}. Note that these correlations are not strictly one for the smaller $\bm{L}$, this is because the $H_m$ is of full column rank in this case.

\section{Performance evaluation}
This section reports the experimental results of the SCNE on two large-scale datasets. 
We first introduce the datasets and explain the heterogeneous feature generation process.
Then we present all the details of the experimental settings and results.
The comparisons between the SCNE and the DNNE are carried out with remarks, and a robustness analysis of the SCNE is discussed at the end.
\subsection{Large-scale datasets}
The SCNE are used to explore two large-scale datasets, which are 
(i) Buzz Prediction on Twitter (Twitter)\footnote{http://ama.liglab.fr/resourcestools/datasets/buzz-prediction-in-social-media/} and (ii) Year Prediction MSD (Year)\footnote{http://archive.ics.uci.edu/ml/datasets/YearPredictionMSD}. 
Table~\ref{Table_Data} shows the samples, features, values and training/testing split. 
These datasets have been analysed previously with some results. 

The Twitter dataset is used for the annotation prediction of the mean number of active discussion (NAD).
This target is a positive integer that describes the popularity of the sample's topic. 
Each sample covers weeks of observation for a specific topic and is described by 77 features. 
All the observations are independent and identically distributed.

Year dataset is used for the prediction of the release year of a song from audio features. 
Songs are mostly western, commercial tracks ranging from 1922 to 2011, with a peak in the year 2000s. 
Among the 90 features, there are 12 belongs to timbre averages and 78 belongs to timbre covariance. 
The first value is the year (target), ranging from 1922 to 2011. 
All the features extracted from the timbre features from The-Echo-Nest-API\footnote{https://developer.spotify.com/spotify-echo-nest-api/}. 
The average and covariance are taken over all segments, each segment being described by a 12-dimensional timbre vector. 

\begin{table}[!h]
\footnotesize 
\centering
\caption{datasets description}
\label{Table_Data}
\begin{tabular}{lcccc}
\hline 
Dataset & Samples & Features & \multicolumn{1}{c}{Values} & Training/Testing \\
\hline 
Buzz Prediction on Twitter & 583,250 & 77 & Numeric &  495,763/87,487 \\ 
Year Prediction MSD & 515,345 & 90 & Numeric &  463,715/51,630\\ 
\hline 
\end{tabular} 
\end{table}

\subsection{Heterogeneous feature generation}
There are various methods to generate the heterogeneous feature sets for the base models. 
In our experiments, the 77 features of Twitter dataset are partitioned into 11 groups based on the introduction of the dataset.
Each group has its specific meaning. 
The details, such as group names, feature indexes and explanations are listed in Table \ref{Table_Twitter_HFGoups}.
\begin{table}[h]
\centering
\tiny 
\caption{Heterogeneous feature groups of Twitter dataset ($M=11$)}\label{Table_Twitter_HFGoups}
\begin{threeparttable}
\begin{tabular}{ccccccccccccc}\hline
Group  &NCD &AI &AS(NA) &BL &NAC &AS(NAC) &CS &AT &NA &ADL &NAD \\
Columns &{[}1,7{]} &{[}8,14{]} &{[}15,21{]} &{[}22,28{]} &{[}29,35{]} &{[}36,42{]} &{[}43,49{]} &{[}50,56{]} &{[}56,63{]} &{[}64,70{]} &{[}71,77{]} \\ \hline
\end{tabular}
\begin{enumerate}
\item NCD: Number of Created Discussions.  This feature measures the number of discussions created at time step t and involving the instance's topic.                                
\item AI: Author Increase. This feature measures the number of new authors interacting on the instance's topic at time t (i.e. its popularity).                                  
\item AS(NA): Attention Level (measured with the number of authors). This feature is a measure of the attention paid to the instance's topic on a social media.                                              
\item BL: Burstiness Level. The burstiness level for a topic z at a time t is defined as the ratio of NCD and NAD.                                                      
\item NAC: Number of Atomic Containers. This feature measures the total number of atomic containers generated through the whole social media on the instance's topic until time t. 
\item AS(NAC): Attention Level (measured with the number of contributions). This feature is a measure of the attention paid to the instance's topic on a social media.                                             
\item CS: Contribution Sparseness. This feature is a measure of spreading of contributions over discussion for the instance's topic at time t.                               
\item AI: Author Interaction. This feature measures the average number of authors interacting on the instance's topic within a discussion.                               
\item NA: Number of Authors. This feature measures the number of authors interacting on the instance's topic at time t.                                                 
\item ADL: Average Discussions Length. This feature directly measures the average length of a discussion belonging to the instance's topic.                                      
\item NAD: Number of Average Discussions. This features measures the number of discussions involving the instance's topic until time t.                                           
\end{enumerate}
\end{threeparttable}
\begin{tabular}{c} \hline
~\hspace{12cm}~
\end{tabular}
\end{table}
According to the data description of the Year dataset, the total 90 features could be partitioned into two groups, timbre averages group (12 features) and timbre covariance (78 features) group.
We equally split the timbre covariance feature group into 6 small groups (TC-a to TC-f).
Thus each small timbre covariance feature group contains 13 features, which is close to the timbre average feature group.
The details could be found in Table \ref{Table_Year_HFGoups}.
\begin{table}[h]
\centering
\tiny 
\caption{Heterogeneous feature groups of Year dataset ($M=7$)} \label{Table_Year_HFGoups}
\begin{threeparttable}
\begin{tabular}{cccccccc}\hline
Group  &TA &TC-a &TC-b &TC-c &TCd &TC-e &TC-f \\
Columns &{[}1,12{]} &{[}13,25{]} &{[}26,38{]} &{[}39,51{]} &{[}52,64{]} &{[}65,77{]} &{[}78,90{]} \\ \hline
\end{tabular}
\begin{enumerate}
\item TA: Timbre Averages. These features are extracted from the 'timbre' features and described by a 12-dimensional timbre vector.                                 
\item TC-[a,\dots,f]: Timbre Covariance. There features are calculated according to the original segments of the songs.
\end{enumerate}
\end{threeparttable}
\begin{tabular}{c} \hline
~\hspace{8cm}~
\end{tabular}
\end{table}
\subsection{Experimental setup} 
\begin{figure}[!h]
\centering
    \begin{subfigure}[b]{0.49\textwidth}
        \includegraphics[width=\textwidth]{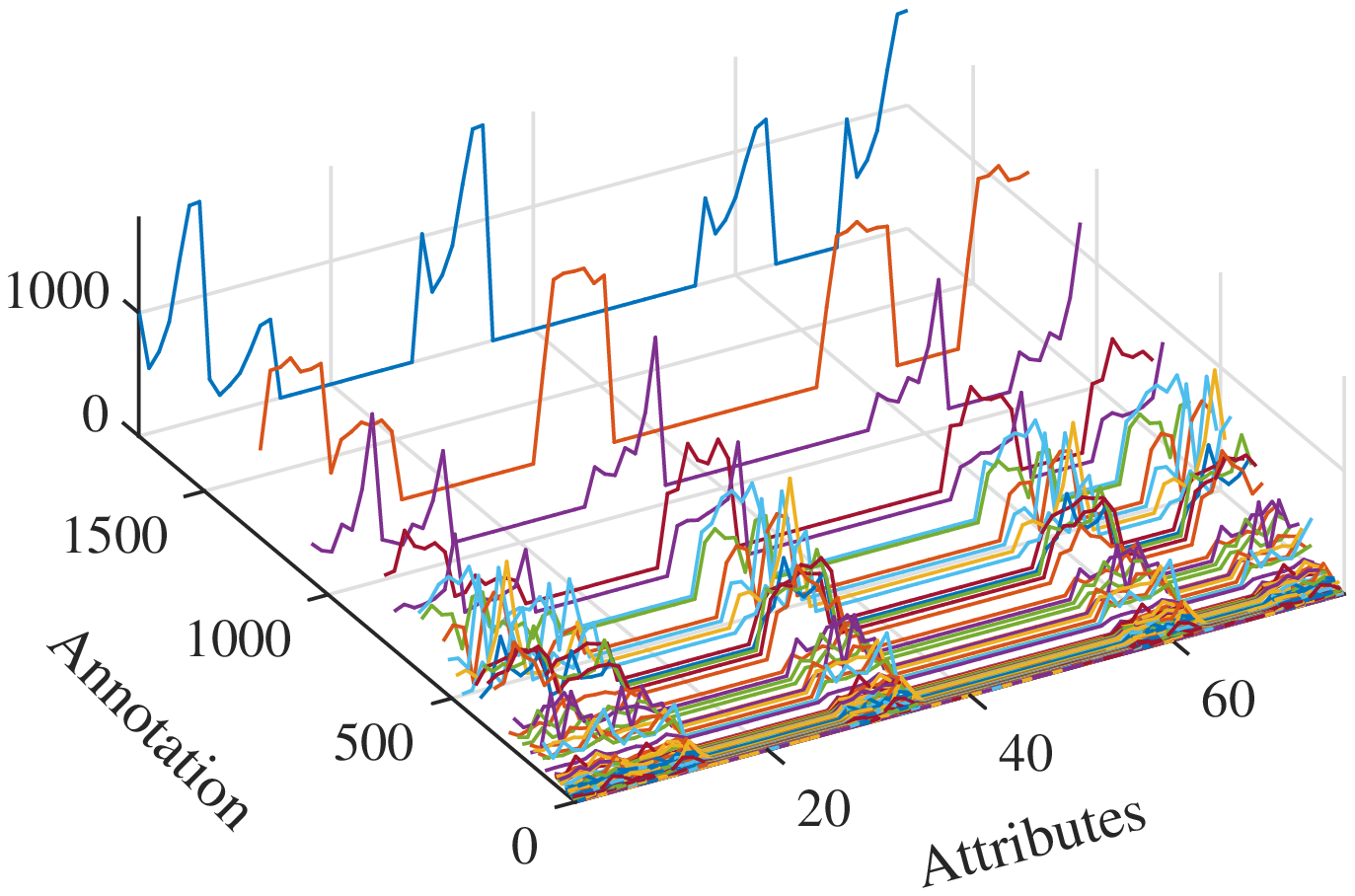}
        \caption{Original.}
        \label{Fig05a: Twitter_Original}    
    \end{subfigure}
    \begin{subfigure}[b]{0.49\textwidth}
    \includegraphics[width=\textwidth]{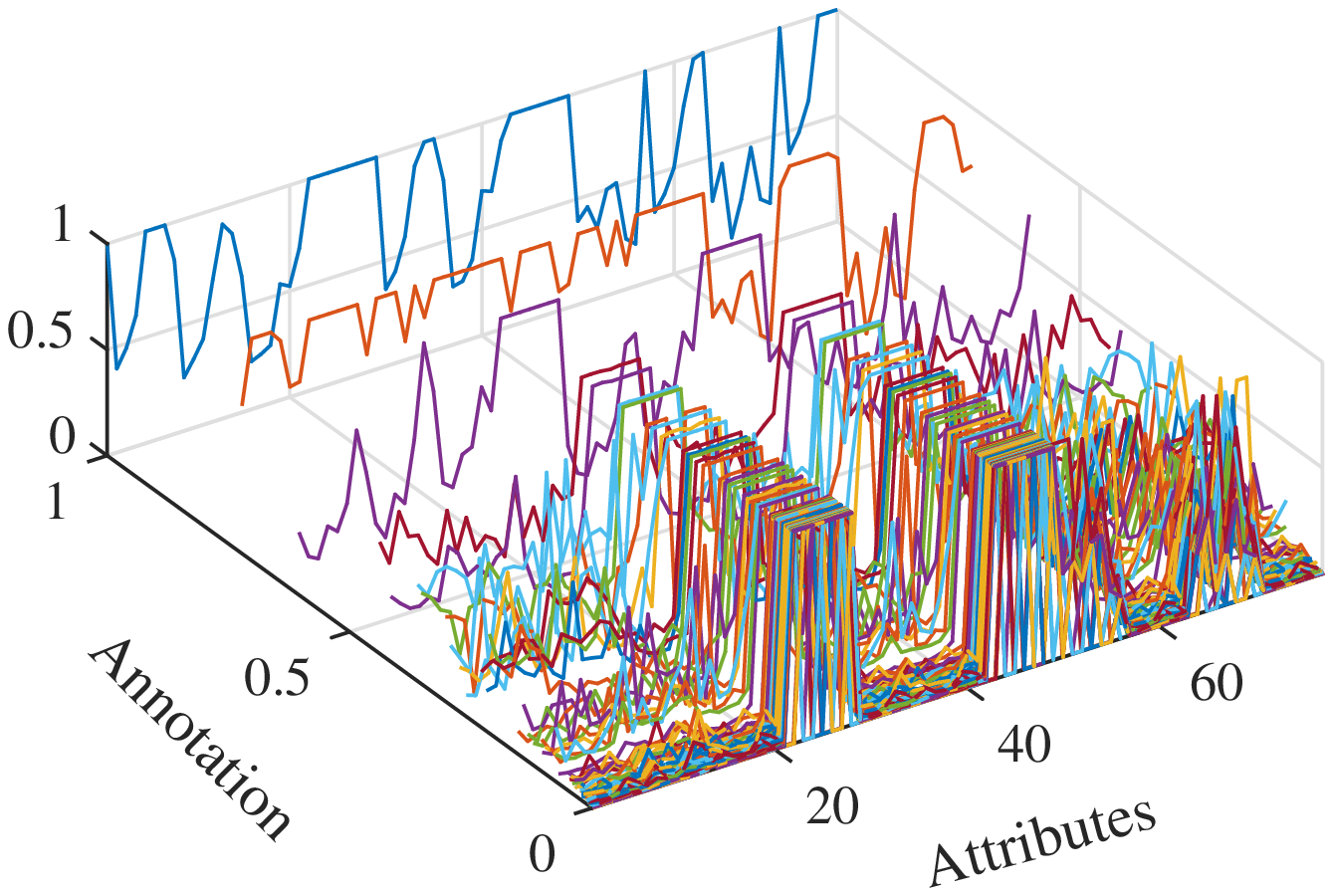}
    \caption{0-1 normalization.}
    \label{Fig05b: Twitter_Zscore}    
    \end{subfigure}    
\caption{The Buzz Prediction on Twitter data: 80 samples from the original dataset. The x-axis represents the attributes, the y-axis shows the Annotation, and the z-axis denotes the values with respect to each attribute. Each line represents one Twitter sample.}
\label{Fig5_Twitter_Data}
\end{figure}

\begin{figure}[!h]
\centering
    \begin{subfigure}[b]{0.49\textwidth}
        \includegraphics[width=\textwidth]{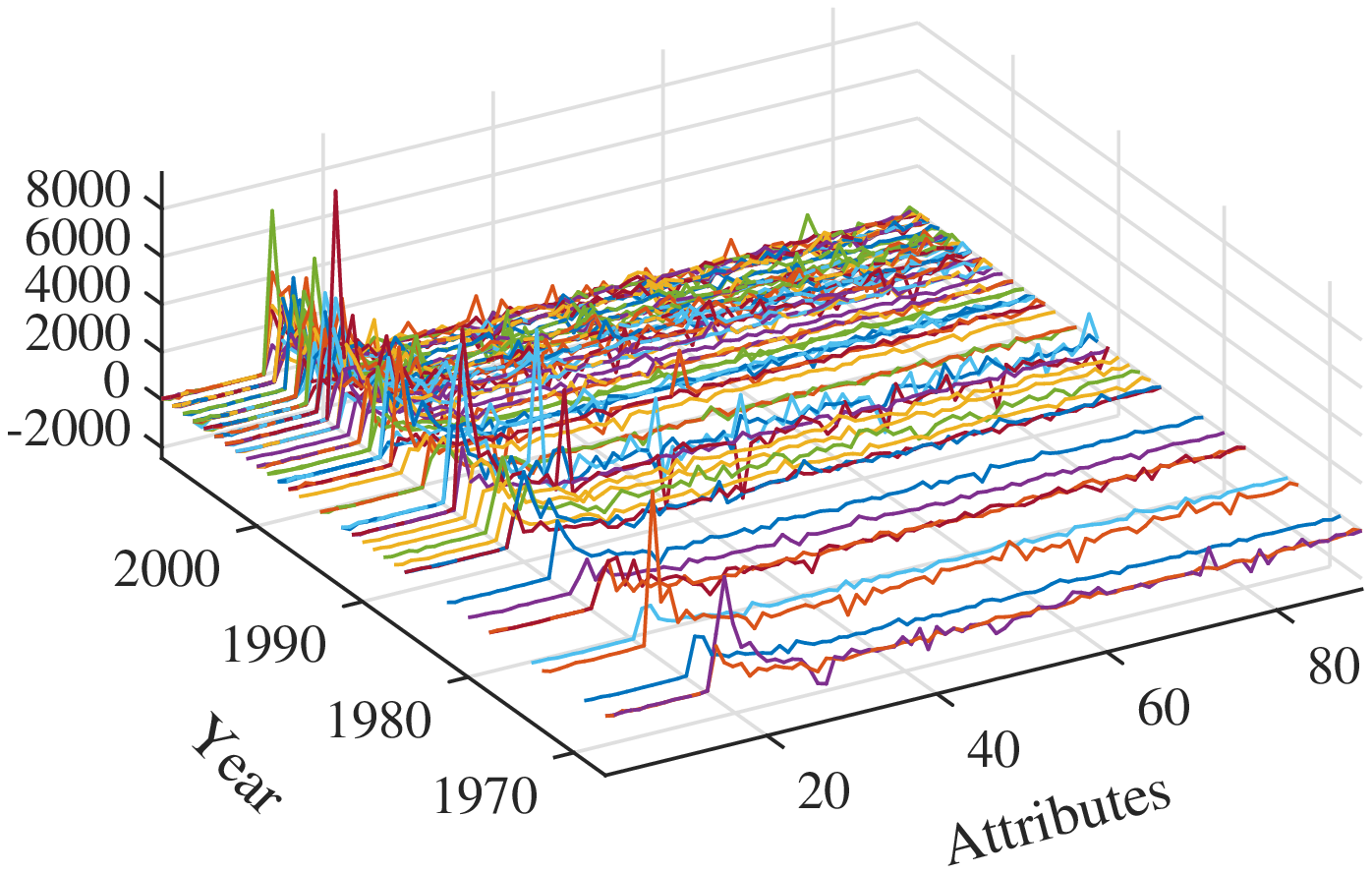}
        \caption{Original.}
        \label{Fig06a: Year_Original}    
    \end{subfigure}    
    \begin{subfigure}[b]{0.49\textwidth}
    \includegraphics[width=\textwidth]{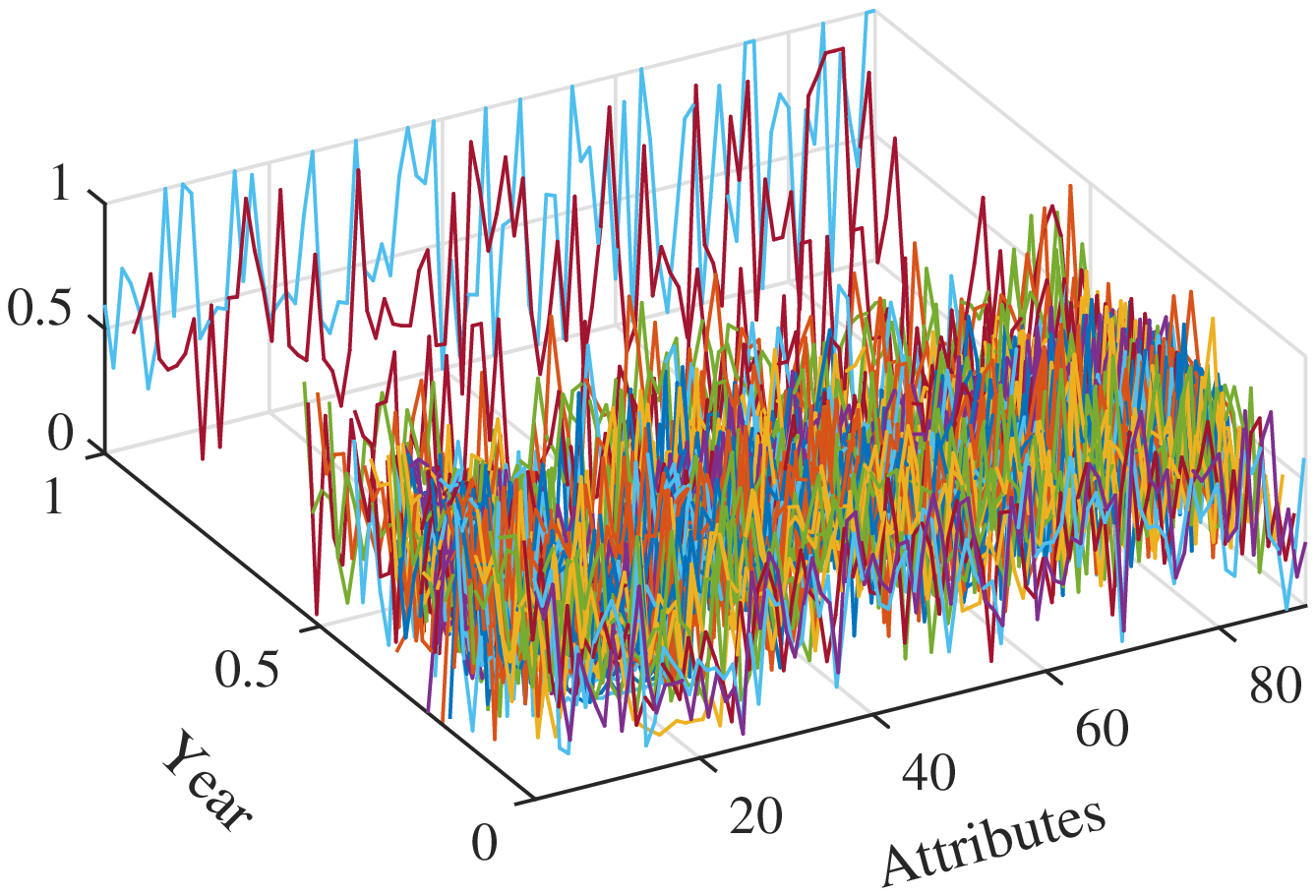}
    \caption{0-1 normalization.}
    \label{Fig06: Year_Zscore}    
    \end{subfigure}    
\caption{The Year Prediction MSD data:  80 samples from the original dataset. The x-axis represents the attributes, the y-axis shows the Year, and the z-axis denotes the timbre values (average and covariance) with respect to each attribute. Each line represents one Year sample.}
\label{Fig6_Year_Data}
\end{figure}

As a common knowledge in machine learning, data preprocessing plays a crucial role before modelling. 
Normalization and standardization are two widely used methods for rescaling data.
The 0-1 normalization could scale all numeric variables into the range $[0,1]$, one possible formula is $x_{new} = (x- x_{min} ) / (x_{max}-x_{min})$. 
The z-score standardization transforms the data to have zero mean and unit variance by the formula $x_{new} = (x-\bar{x})/ \sigma $, which indicates how many standard deviations an element is from the mean. However, if there were outliers in the dataset, normalization will certainly scale the “normal” data to a very small interval. Our experiments aim to demonstrate the capability of the SCNE for large-scale datasets, here  we choose  the 0-1 normalization method for data preprocessing and assume there is no outlier in the dataset. To show the data distribution, we randomly select a small batch of samples from each dataset and present them in Figs.~\ref{Fig5_Twitter_Data} and \ref{Fig6_Year_Data}, respectively.

\begin{figure}[h]
\centering
\includegraphics[scale= 0.8]{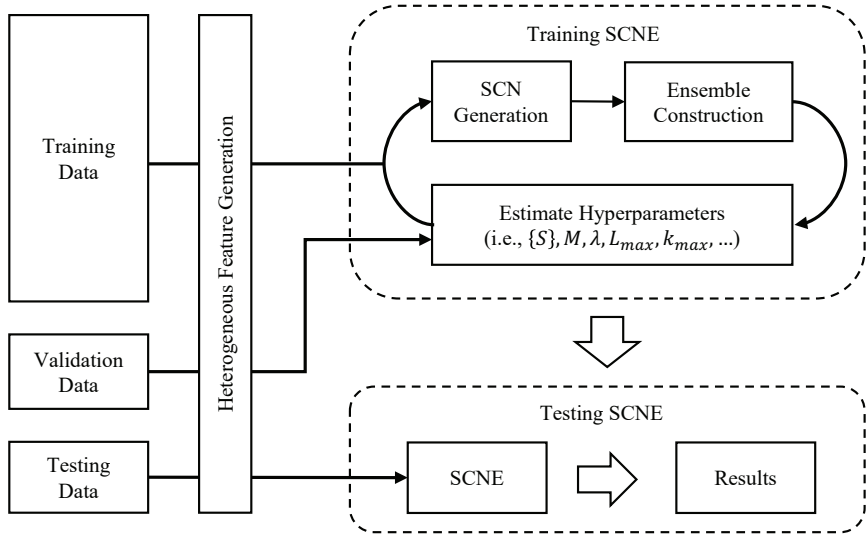}
\caption{ Experimental diagram of the SCNE including the SCN generation and ensemble construction.}
\label{Fig7_ExperimentalDiagram}
\end{figure}

All the experiments are designed, repeated and followed the same procedure. 
Fig.~\ref{Fig7_ExperimentalDiagram} presents the general experimental diagram. 
The arrows indicate the direction of the data feeds.
The experiments are designed in two stages, training and testing, indicated by the dash-lined box in the diagram.
In the training stage, the training dataset is used to build the SCNE, and the validation data is used to adjust and refine the hyper-parameters of the ensemble, such as $\{S\}$, $M$, $\lambda$, $L_{max}$ and $k_{max}$ (In our experiments, the heterogeneous feature set $\{S\}$ and the base model number $M$ is predefined). 
When all these parameters are properly estimated, the training and validation data will be used together as one combined set to retrain the SCNE again; then this final ensemble will be tested on the testing dataset.
The testing results could be seen as a good indicator for the generalization ability of the SCNE.
Unlike some benchmark dataset, there is no ready-made training or testing sets of Twitter data, which leaves us various choices to partition the data.
For Twitter data, we apply $70\%$ of the total samples for training, $15\%$ for validation and the rest $15\%$ for testing, denoted as ``70-15-15".
The Year dataset has been specified with the testing data (almost $10\%$), so we randomly split the rest $90\%$ samples into two parts $70\%$ for training and $20\%$ for validation, denoted as ``70-20-10". 

\subsection{Results and discussion}
The following section presents all the training, validation, testing results and parameter estimations of SCNE on Twitter and Year datasets. 
All the experimental conditions and details are stated and specified as follows for the convenience to repeat the experiments.
The comparative ensemble is the DNNE with the same structure of the SCNE.

For large datasets, such as the Twitter and Year in this paper, the general training-validation procedure for estimating hyper-parameters or preventing model overfitting usually costs a large amount of time. 
According to the SC-III algorithm, each base model needs to compute the results of all its training inputs, and then evaluate all the candidate nodes to add one or a few hidden nodes in one SC-search loop. 
This would take more time to estimate the proper hidden node number $L_m$ of each SCN models of the ensemble.
To accelerate this procedure, a down-sampling approach is applied, that is, instead of using all the samples, we randomly select two sub-datasets to estimate the parameters from the training and validation datasets, respectively,  and repeat this procedure for several times to obtain statistic results.
In this paper, we choose 70,000 samples from the training set and 30,000 from the validation set with heterogeneous features as the training-validation datasets for the estimation of the hidden node number of SCNs in the ensemble. 
Then build each SCN model with increasing hidden nodes to find the hidden node number with respect to the minimal validation error.
The search range for the $L$  of each SCN model is $[1, 120]$. 

\begin{figure}[!h]
\centering
    \begin{subfigure}[b]{0.49\textwidth}
        \includegraphics[width=\textwidth]{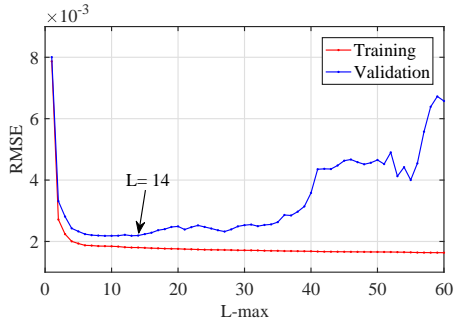}
        \caption{Twitter, m =1, index = 4, $L_m = 14$.}
        \label{}    
    \end{subfigure}
    \begin{subfigure}[b]{0.49\textwidth}
        \includegraphics[width=\textwidth]{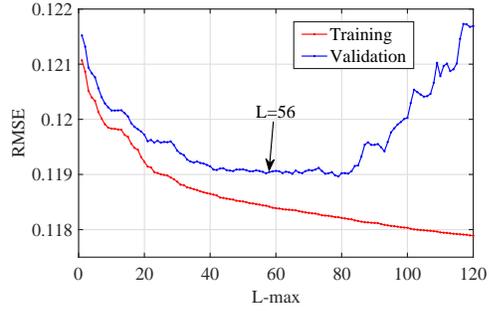}
        \caption{Year, m =2, index = 3, $L_m = 56$.}
        \label{}    
    \end{subfigure}      
\caption{Examples of the training-validation error lines for estimating the hidden nodes of the SCN model in the ensemble on the Twitter and Year datasets. (a) SCN base model 1, index = 4, on the Twitter dataset; (b) SCN base model 2, index = 3, on the Year dataset. For better visual effect, the axis of (a) has been adjusted from $[0, 120]$ to $[0, 60]$.}
\label{Fig07}
\end{figure}

Fig.~\ref{Fig07} presents two examples of the training-validation error lines for the hidden nodes estimations of the base models on Twitter and Year datasets, respectively. This procedure is repeated for 10 times. The estimated hidden node numbers of the SCNs are summarised in Tables~\ref{Table_SCNE1} and \ref{Table_SCNE2}. The medium of each column is chosen as the hidden node number for each SCN base model.

For fair comparisons, the DNNE shares the same architecture as the SCNE, that is, each RVFL base model has the same number of hidden nodes as the corresponding SCN base model.
According to \cite{li2017insights}, the range of the random weights and biases of the RVFL networks, $[-\alpha, +\alpha]$, needs to be estimated to improve the training performance. 
In our experiments, we compare the RVFL models with varying $\alpha$ from $0.5$ to $1.4$.
The results are listed in Tables~\ref{Talbe_RVFLE1} and \ref{Talbe_RVFLE2}. 
The training data and validation data are sampled as the same as those for the SCN model.
Each result is the average of 10 independent repeats with the same $\alpha$.
For each RVFL base model, the $\alpha^*$ with the minimal validation error is chosen for the RVFL base model used to construct the final DNNE. 
\begin{table}[H]
\footnotesize 
\centering
\caption{SCNE base model hidden node estimations on the Twitter dataset}
\label{Table_SCNE1}
\begin{tabular}{c|cccccccccccc}\hline
\multicolumn{12}{c}{SCNE Base Model (Twitter)} \\\hline
\multicolumn{1}{c|}{Index $\setminus m$} & \multicolumn{1}{c}{1} & \multicolumn{1}{c}{2} & \multicolumn{1}{c}{3} & \multicolumn{1}{c}{4} & \multicolumn{1}{c}{5} & \multicolumn{1}{c}{6} & \multicolumn{1}{c}{7} & \multicolumn{1}{c}{8} & \multicolumn{1}{c}{9} & \multicolumn{1}{c}{10} & \multicolumn{1}{c}{11} \\ \hline
1       & 17   & 14 & 44   & 66   & 10 & 9  & 42 & 41 & 23   & 30   & 15   \\
2       & 19   & 22 & 14   & 63   & 25 & 10 & 45 & 48 & 14   & 17   & 13   \\
3       & 10   & 20 & 23   & 61   & 18 & 32 & 45 & 33 & 21   & 18   & 17   \\
4       & 14   & 30 & 25   & 77   & 10 & 18 & 40 & 24 & 24   & 29   & 19   \\
5       & 23   & 32 & 15   & 64   & 17 & 15 & 41 & 31 & 18   & 39   & 16   \\
6       & 16   & 30 & 24   & 69   & 23 & 21 & 39 & 39 & 22   & 24   & 10   \\
7       & 17   & 17 & 15   & 90   & 17 & 6  & 38 & 34 & 32   & 23   & 19   \\
8       & 14   & 30 & 12   & 53   & 16 & 13 & 39 & 41 & 29   & 32   & 14   \\
9       & 17   & 18 & 30   & 72   & 23 & 13 & 41 & 33 & 18   & 30   & 17   \\
10      & 7    & 13 & 20   & 73   & 15 & 31 & 43 & 36 & 14   & 20   & 19   \\ \hline
Medium  & 17 & 21 & 22 & 68 & 17 & 14 & 41 & 35 & 22 & 27 & 17 \\\hline
\end{tabular}
\end{table}
\begin{table}[H]
\footnotesize 
\centering
\caption{SCNE base model hidden node estimations on the Year dataset}
\label{Table_SCNE2}
\begin{tabular}{c|ccccccc}\hline
 \multicolumn{8}{c}{SCNE Base Model (Year)} \\ \hline
\multicolumn{1}{c|}{Index $\setminus m$} & \multicolumn{1}{c}{1} & \multicolumn{1}{c}{2} & \multicolumn{1}{c}{3} & \multicolumn{1}{c}{4} & \multicolumn{1}{c}{5} & \multicolumn{1}{c}{6} & \multicolumn{1}{c}{7} \\ \hline
1       & 70 & 23 & 17   & 36  & 51 & 15 & 18    \\
2       & 61 & 54 & 32   & 19  & 17 & 21 & 18    \\
3       & 62 & 56 & 17   & 29  & 36 & 29 & 37    \\
4       & 60 & 44 & 28   & 23  & 43 & 26 & 30    \\
5       & 61 & 22 & 23   & 20  & 22 & 18 & 33    \\
6       & 61 & 24 & 19   & 24  & 32 & 21 & 18    \\
7       & 60 & 21 & 28   & 20  & 38 & 22 & 33    \\
8       & 65 & 28 & 26   & 19  & 18 & 19 & 38    \\
9       & 40 & 27 & 21   & 50  & 16 & 22 & 23    \\
10      & 60 & 27 & 28   & 19  & 12 & 29 & 20    \\ \hline
Medium  & 61 & 27 & 25   & 22  & 27 & 22 & 27    \\ \hline
\end{tabular}
\end{table}
The following results show the advantages of the SCNE on Twitter and Year datasets.
When the hidden node numbers of the SCNE and random weights ranges of the DNNE are estimated according to the previous experiments, we retrain the SCNE and DNNE with the full training datasets with the regularizing factor $\lambda = 0.10$ and the iteration number $k_{max} = 10$. 
Tables~\ref{Table08} and~\ref{Table09} present the final training and testing results (i.e., the root mean square error, RMSE) of SCNE and DNNE with respect to 4 construction algorithms. 
All the results are the averages of 10 independent repeats.
It is clear to see that the SCNE has lower training and testing RMSE than the DNNE on both datasets.
The pseudo-inverse method and iterative schemes got much lower errors than the naive algorithm. 
Despite the training time cost, these results clearly indicate that the proposed SCNE works better than the DNNE.
\begin{table}[H]
\footnotesize 
\centering
\caption{RVFL base model random weight range estimations for the Twitter dataset}
\label{Talbe_RVFLE1}
\begin{tabular}{c|ccccccccccc}\hline
  \multicolumn{12}{c}{Validation RMSE on Twitter data($10^{-3}$)} \\\hline
\multicolumn{1}{c|}{$\alpha \setminus m$} & \multicolumn{1}{c}{1} & \multicolumn{1}{c}{2} & \multicolumn{1}{c}{3} & \multicolumn{1}{c}{4} & \multicolumn{1}{c}{5} & \multicolumn{1}{c}{6} & \multicolumn{1}{c}{7} & \multicolumn{1}{c}{8} & \multicolumn{1}{c}{9} & \multicolumn{1}{c}{10} & \multicolumn{1}{c}{11} \\ \hline
0.5 & 4.244 & 9.574  & 6.014 & 7.652 & 2.226 & 3.756 & 8.012 & 9.821  & 4.448 & 8.213 & 2.108 \\
0.6 & 2.527 & 14.557 & 5.060 & 7.842 & 3.143 & 4.242 & 7.845 & 8.590  & 3.524 & 8.578 & 2.484 \\
0.7 & 3.990 & 4.922  & 7.852 & 8.085 & 2.849 & 4.141 & 7.749 & 10.030 & 3.990 & 8.146 & 2.949 \\
0.8 & 2.117 & 6.759  & 4.602 & 8.206 & 2.731 & 3.744 & 8.048 & 8.550  & 4.639 & 8.596 & 2.580 \\
0.9 & 2.133 & 4.849  & 5.584 & 8.011 & 2.332 & 3.964 & 7.899 & 8.375  & 5.041 & 8.553 & 2.198 \\
1.0 & 2.064 & 4.001  & 3.899 & 8.206 & 2.332 & 3.823 & 7.872 & 8.336  & 5.082 & 8.468 & 2.297 \\
1.1 & 2.568 & 9.704  & 4.361 & 8.268 & 3.029 & 4.069 & 8.218 & 9.124  & 5.293 & 7.915 & 2.593 \\
1.2 & 2.666 & 4.211  & 3.992 & 7.904 & 2.308 & 4.289 & 7.792 & 11.154 & 5.207 & 8.059 & 2.337 \\
1.3 & 2.509 & 5.262  & 4.551 & 8.073 & 2.137 & 4.259 & 7.719 & 8.250  & 3.602 & 8.527 & 2.295 \\
1.4 & 2.325 & 5.375  & 5.131 & 8.161 & 2.418 & 3.505 & 7.973 & 8.389  & 3.713 & 8.076 & 2.327 \\\hline
$\alpha^*$& 1.0 & 1.0 & 1.0 & 0.5 & 1.3 & 1.4 & 1.3 & 1.3 & 0.6 & 1.1 & 0.5  \\ \hline
\end{tabular}
\end{table}
\begin{table}[H]
\footnotesize 
\centering
\caption{RVFL base model random weight range estimations for the Year dataset}
\label{Talbe_RVFLE2}
\begin{tabular}{c|ccccccc}\hline
\multicolumn{8}{c}{Validation RMSE on Year data ($10^{-1}$)} \\ \hline
\multicolumn{1}{c|}{$\alpha \setminus m$} & \multicolumn{1}{c}{1} & \multicolumn{1}{c}{2} & \multicolumn{1}{c}{3} & \multicolumn{1}{c}{4} & \multicolumn{1}{c}{5} & \multicolumn{1}{c}{6} & \multicolumn{1}{c}{7}\\ \hline
0.5 & 1.101 & 1.202 & 1.206 & 1.202 & 1.195 & 1.204 & 1.208 \\
0.6 & 1.103 & 1.209 & 1.207 & 1.201 & 1.197 & 1.204 & 1.209 \\
0.7 & 1.103 & 1.204 & 1.207 & 1.200 & 1.196 & 1.207 & 1.209 \\
0.8 & 1.105 & 1.202 & 1.206 & 1.202 & 1.198 & 1.202 & 1.210 \\
0.9 & 1.105 & 1.203 & 1.206 & 1.200 & 1.197 & 1.207 & 1.211 \\
1.0 & 1.103 & 1.204 & 1.213 & 1.201 & 1.200 & 1.207 & 1.211 \\
1.1 & 1.105 & 1.204 & 1.213 & 1.200 & 1.194 & 1.202 & 1.215 \\
1.2 & 1.102 & 1.204 & 1.207 & 1.203 & 1.199 & 1.202 & 1.213 \\
1.3 & 1.100 & 1.207 & 1.206 & 1.202 & 1.193 & 1.208 & 1.211 \\
1.4 & 1.101 & 1.204 & 1.207 & 1.197 & 1.198 & 1.206 & 1.212 \\\hline
$\alpha^*$ & 1.3   & 0.8   & 0.8   & 1.4   & 1.3   & 0.8   & 0.5  \\\hline
\end{tabular}
\end{table}
\begin{table}[H]
\footnotesize 
\centering
\caption{Twitter dataset ensemble results ($M =11$)}
\label{Table08}
\begin{tabular}{l|lcc}
\hline
Ensemble Type                  & Algorithm         & Training RMSE($10^{-3}$) & Test RMSE($10^{-3}$) \\  \hline
\multirow{4}{*}{SCNE}  & Naive             & 3.950$\pm$0.0037   & 3.678$\pm$0.0041  \\
                               & Pseudo-inverse    & 3.811$\pm$0.0032   & 3.542$\pm$0.0031  \\
                               & Block Jacobi      & 3.811$\pm$0.0032   & 3.542$\pm$0.0042  \\
                               & Block Gauss-Seidel & 3.811$\pm$0.0031   & 3.542$\pm$0.0041  \\ \hline
\multirow{4}{*}{DNNE} & Naive             & 3.959$\pm$0.0036   & 3.839$\pm$0.0114  \\  
                               & Pseudo-inverse    & 3.829$\pm$0.0050   & 3.696$\pm$0.0104  \\
                               & Block Jacobi      & 3.824$\pm$0.0052   & 3.694$\pm$0.0107  \\
                               & Block Gauss-Seidel & 3.824$\pm$0.0052   & 3.694$\pm$0.0101  \\ \hline
\end{tabular}
\end{table}
\begin{table}[H]
\footnotesize 
\centering
\caption{Year dataset ensemble results ($M =7$)}
\label{Table09}
\begin{tabular}{l|lcc}
\hline
Ensemble Type             & Algorithm & Training RMSE($10^{-1}$) & Test RMSE($10^{-1}$) \\
\hline
\multirow{4}{*}{SCNE}  & Naive             & 1.16351$\pm$0.00101  & 1.14829$\pm$0.00096 \\
                               & Pseudo-inverse    & 1.16093$\pm$0.00165  & 1.14720$\pm$0.00121 \\
                               & Block Jacobi      & 1.16042$\pm$0.00083  & 1.14690$\pm$0.00114 \\
                               & Block Gauss-Seidel & 1.16042$\pm$0.00097  & 1.14690$\pm$0.00109 \\\hline
\multirow{4}{*}{DNNE} & Naive             & 1.16451$\pm$0.00097  & 1.16129$\pm$0.00098 \\
                               & Pseudo-inverse    & 1.16122$\pm$0.00113  & 1.15320$\pm$0.00115 \\
                               & Block Jacobi      & 1.16122$\pm$0.00114  & 1.15320$\pm$0.00115 \\
                               & Block Gauss-Seidel & 1.16122$\pm$0.00103  & 1.15320$\pm$0.00113 \\
\hline
\end{tabular}
\end{table}
\begin{figure}[H]
\centering
    \begin{subfigure}[b]{0.47\textwidth}
    \centering
        \includegraphics[width=0.9\textwidth]{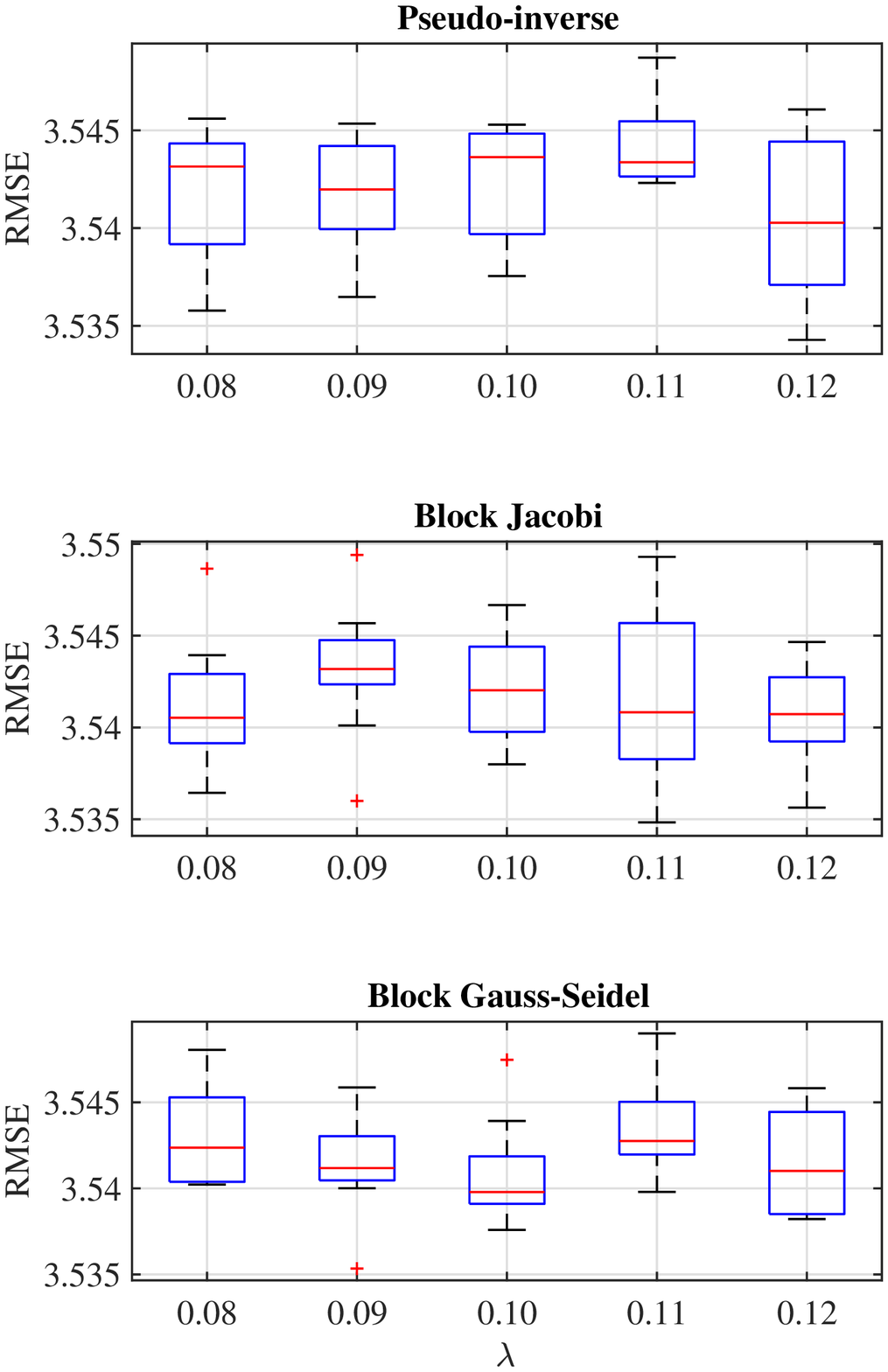}
        \caption{Twitter dataset.}
        \label{}    
    \end{subfigure}
    \begin{subfigure}[b]{0.47\textwidth}
    \centering
        \includegraphics[width=0.9\textwidth]{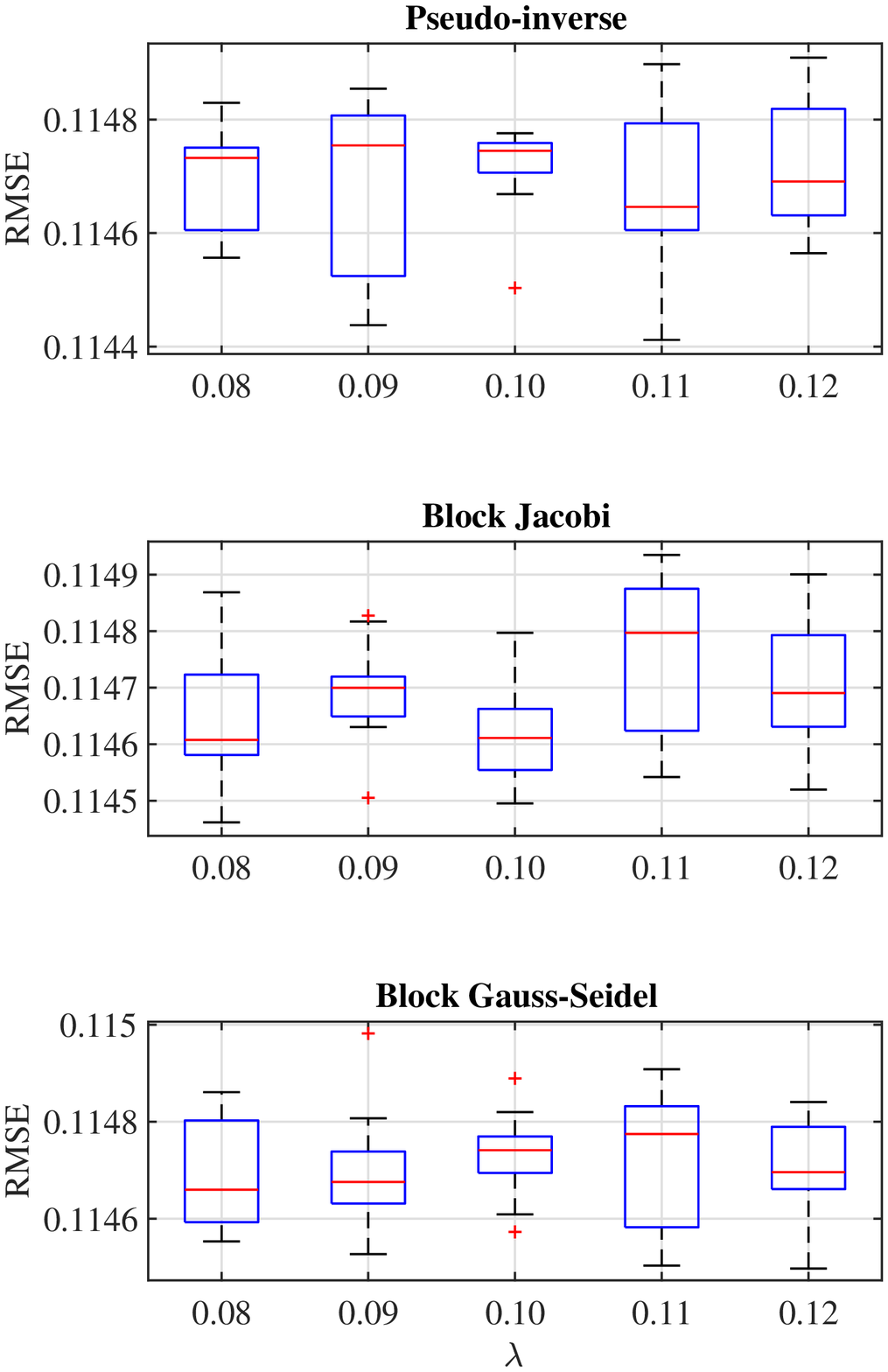}
        \caption{Year dataset.}
        \label{} 
    \end{subfigure}          
\caption{Distributions of the SCNE testing results for the robustness analysis on Twitter and Year datasets with $\lambda$ from 0.08 to 0.12. The figures should be scaled by $10^{-3}$ and $10^{-1}$ for the Twitter and Year datasets, respectively.}
\label{Fig09}
\end{figure}
\begin{remark} 
In SCNE construction, it should be emphasized that the pseudo-inverse method costs much more computing time and memory than the iterative methods with a large $\bm{L}$.
Ideally, applying the iterative methods, block Jacobi and block Gauss-Seidel methods, could lower the complexity to approximate $O(\sum_{m=1}^{M} L_m^3)$.
In the Twitter experiments, the peak of the memory cost of the pseudo-inverse method is about 7.9GB, which is the limit of our computer (8GB RAM) and it tends to take more. 
However, for iterative methods, the peaks are at almost the same level as the naive method, never beyond 5.4GB, which means on the same machine, the iterative methods have more potentials than the pseudo-inverse method for large-scale data modelling.
\end{remark}  

\subsection{Robustness analysis}
The regularizing factor $\lambda = 0.10$ is used for the construction of SCNE in previous experiments. 
To investigate the reliability of the proposed SCNE system, an  analysis on the robustness of the modelling performance with respect to some key learning  parameters should be carried out. In this work, we focus on the robustness analysis for the regularizing factor $\lambda$ used in the NCL. 
The hidden nodes and data partitions for SCN base models are the same as the previous experiments.
Only the value of the $\lambda$ is changed here.
Fig.~\ref{Fig09} depicts the distribution of 10 testing results of the SCNE in box-plots with respect to  $\lambda = [0.08, 0.09, 0.10, 0.11, 0.12]$.
On each box, the central red mark indicates the median, and the bottom and top edges of the box indicate the first and third percentiles ($Q1$ and $Q3$), respectively.  
The top and bottom bars indicate minimum and maximum within the inner fence $[Q1-1.5\times IQR, Q3+1.5\times IQR ]$ where $IQR = Q3-Q1$.  
The values beyond the inner fence are suspected outliers plotted individually using the red `+' symbol.
For both datasets, the results show that the $\lambda$ affects the SCNE testing performance slightly but not significantly within a certain range.
 
\section{Conclusion}
Analysing more data quickly with higher accuracy turns to be significant nowadays because of a vast number of real-world applications from various domains.  Traditional machine learning techniques, such as neural networks with optimization-based learning algorithms, support vector machines and decision trees, are hardly applied for large-scale datasets.  Ensemble learning with its theoretical framework helps in  improving the generalization performance of the base learner models, but it still has some limitations on the efficiency and scalability for dealing with large-scale data modelling problems. 

This paper contributes to the development of randomized neuro-ensemble with heterogeneous features, where the stochastic configuration networks are employed as base learners and the well-known negative correlation learning strategy is adopted to evaluate the output weights of the SCNE model. To overcome the challenge in computing a pseudo-inverse of a huge sized linear equation system used in the least squares method, we suggest to utilize the block Jacobi and block Gauss-Seidel iterative schemes for problem solving. Some analyses and discussions on these solutions for evaluating the output weights are given by a demonstration. Simulation results clearly indicate that it is necessary to apply the ridge regression method for building the SCN base models, so that the resulting SCNE models with the iterative schemes can be consistent with the one built by using the non-iterative method in terms of the correlationship of the output weights. 

The reported results in the demonstration implies  that the statement on the speediness of the pseudo-inverse-based solution for building randomized learner models (either for single or ensemble models) is valid only for smaller datasets. Indeed, its computational complexity and time cost are very high, even infeasible, for large-scale datasets.  Experimental results with comparisons over two large-scale benchmark datasets show that the proposed SCNE always outperforms the DNNE. Robustness analysis on the modelling performance with respect to the regularizing factor used in NCL reveals that our proposed ensemble system performs robustly with  good potential for large-scale data analytics. 
Further researches on improved feature grouping methodology, robust large-scale data regression \cite{WangandLi_RSCN} and enhancement of the generalization performance of the ensemble system are being expected.

\bibliographystyle{elsarticle-num}

\begin{thebibliography}{}
\bibitem{alhamdoosh2014fast}
M.~Alhamdoosh, D.~Wang, 
\newblock Fast decorrelated neural network ensembles with random weights,
\newblock {Inf. Sci.} 264 (2014) 104--117.
\bibitem{breiman1996bagging}
L.~Breiman,
\newblock Bagging predictors,
\newblock {Machine Learning} 24 (2) (1996) 123--140.
\bibitem{breiman2001random}
L.~Breiman,
\newblock Random forests,
\newblock {Machine Learning} 45 (1) (2001) 5--32.
\bibitem{brown2005managing}
G.~Brown, J.~L.~Wyatt, P.~Ti{\v{n}}o,
\newblock Managing diversity in regression ensembles,
\newblock {J. Mach. Learn. Res.} 6 (2005) 1621--1650.
\bibitem{chen2009regularized}
H.~Chen, X.~Yao,
\newblock Regularized negative correlation learning for neural network
  ensembles,
\newblock {\em IEEE Trans. Neural Networks} 20 (12) (2009) 1962--1979.
\bibitem{cui2016high}
C.~Cui, D.~Wang,
\newblock High dimensional data regression using lasso model and neural networks with random weights,
\newblock {Inf. Sci.} 372 (2016) 505--517.
\bibitem{geman1992neural}
S.~Geman, B~Elie, D.~Ren{\'e},
\newblock Neural networks and the bias/variance dilemma,
\newblock {Neural Computation} 4 (1) (1992) 1--58.
\bibitem{golub1996Matrix}
G.~H.~Golub, C.~F.~Van Loan,
\newblock Matrix Computations. Third edition. 
\newblock {The Johns Hopkins University Press}, Baltimore, 1996. 
\bibitem{hansen1990neural}
L.~K.~Hansen, P.~Salamon,
\newblock Neural network ensembles,
\newblock {IEEE Trans. on Pattern Anal. Mach. Intell.} 12 (1990) 993--1001.
\bibitem{hinton2006reducing}
G.~E.~Hinton, R.~R~Salakhutdinov,
\newblock Reducing the dimensionality of data with neural networks,
\newblock {Science}  313 (2006) 504--507.
\bibitem{igelnik1995stochastic}
B.~Igelnik, Y.~H.~Pao,
\newblock Stochastic choice of basis functions in adaptive function approximation and the functional-link net,
\newblock {IEEE Trans. Neural Networks} 6 (6) (1995) 1320--1329.
\bibitem{igelnik1999ensemble}
B.~Igelnik, Y.~H.~Pao, S.~R.~LeClair, C.~Y.~Shen,
\newblock The ensemble approach to neural-network learning and generalization,
\newblock {IEEE Trans. Neural Networks} 10 (1) (1999) 19--30.
\bibitem{jordan2015machine}
M.~Jordan, T.~Mitchell,
\newblock Machine learning: Trends, perspectives, and prospects,
\newblock {Science} 349 (6245) (2015) 255--260.
\bibitem{li2017insights}
M.~Li, D.~Wang,
\newblock Insights into randomized algorithms for neural networks: Practical issues and common pitfalls,
\newblock {Inf. Sci.} 382--383 (2017) 170-178.
\bibitem{li2015multisource}
W.~T.~Li, D.~Wang, T.~Y.~Chai,
\newblock Multi-source data ensemble modeling for clinker free lime content in rotary kiln sintering processes,
\newblock {IEEE Trans. Systems, Man and Cybernetics: Systems} 45 (2) (2015) 303-314.
\bibitem{liu1999ensemble}
Y.~Liu, X.~Yao,
\newblock Ensemble learning via negative correlation,
\newblock {IEEE Trans. Neural Networks} 12 (10) (1999) 1399--1404.
\bibitem{lopez2013randomized}
D.~Lopez-Paz, P.~Hennig, B.~Sch{\"o}lkopf,
\newblock The randomized dependence coefficient,
\newblock In {Advances in Neural Information Processing Systems} (2013) 1--9.
\bibitem{nguyen2014multivariate}
H.~V. Nguyen, E.~M{\"u}ller, J.~Vreeken, P.~Efros,  K.~B{\"o}hm.
\newblock Multivariate maximal correlation analysis,
\newblock  in: {Proceeding of International Conference on Machine Learning}, 2014, pp. 775--783.
\bibitem{pao1992functional}
Y.~H. Pao, Y.~Takefji,
\newblock Functional-link net computing,
\newblock {IEEE Comput. J.} 25 (5) (1992) 76--79.
\bibitem{peng2005feature}
H.~Peng, F.~Long, C.~Ding.
\newblock Feature selection based on mutual information criteria of max-dependency, max-relevance, and min-redundancy,
\newblock {IEEE Trans. Pattern Analy. Mach. Intell.} 27 (8) (2005) 1226--1238.
\bibitem{polikar2012ensemble}
C.~ Zhang, Y.~ Ma,
\newblock {Ensemble Machine learning}, Springer US, 2012.
\bibitem{reshef2011detecting}
D.~N. Reshef, Y.~A. Reshef, H.~K. Finucane, S.~R. Grossman, G.~McVean, P.~J. Turnbaugh, E.~S. Lander, M.~Mitzenmacher, P.~C. Sabeti,
\newblock Detecting novel associations in large data sets,
\newblock {Science} 334 (6062) (2011) 1518--1524.
\bibitem{rosen1996ensemble}
B.~E.~Rosen, 
\newblock Ensemble learning using decorrelated neural networks,
\newblock {Connection Science} 8 (3-4) (1996) 373--384.
\bibitem{scardapane2015distributed}
S.~Scardapane,  D.~Wang,  P.~Massimo,  U.~Aurelio,
\newblock Distributed learning for random vector functional-link networks,
\newblock {Inf. Sci.} 301 (2015) 271--284.
\bibitem{scardapane2016decentralized}
S.~Scardapane,  D.~Wang,  P.~Massimo,
\newblock A decentralized training algorithm for echo state networks in distributed big data applications,
\newblock {Neural Networks} 78 (2016) 65-74.
\bibitem{schapire1990strength}
R.~E.~Schapire,
\newblock The strength of weak learnability,
\newblock {Machine Learning} 5 (2) (1990) 197--227.
\bibitem{ueda1996generalization}
N.~Ueda, R.~Nakano,
\newblock Generalization error of ensemble estimators,
\newblock in: Proceedings of IEEE Conference on Neural Networks, Vol.1. 1992, pp. 90--95.
\bibitem{WangandLi_SCN}
D.~Wang, M.~Li,
\newblock Stochastic configuration networks: Fundamentals and algorithms,
\newblock {arXiv:1702.03180} [cs.NE], (2017).
\bibitem{WangandLi_RSCN}
D.~Wang, M.~Li,
\newblock Robust stochastic configuration networks with kernel density estimation for uncertain data regression,
\newblock {Inf. Sci.} 412-–413 (2017) 210--222.
\bibitem{young1971Iterative}
D.~M.~Young,
\newblock Iterative Solution of Large Linear Systems,
\newblock {Academic Press}, New York, 1971. 

\end{thebibliography}

\end{document}